\documentclass[lettersize,journal]{IEEEtran}
\usepackage{amsmath,amssymb,amsfonts,amsthm}
\usepackage{mathtools}
\usepackage{bm}
\usepackage{nicefrac}
\usepackage{algorithm}
\usepackage{algpseudocode}
\usepackage{graphicx}
\usepackage[caption=false,font=normalsize,labelfont=sf,textfont=sf]{subfig}
\usepackage{booktabs}
\usepackage{array}
\usepackage{tabularx}
\usepackage{multirow}
\usepackage{makecell}
\usepackage[table]{xcolor}
\usepackage{textcomp}
\usepackage{microtype}
\usepackage{url}
\usepackage{verbatim}
\usepackage{cite}
\usepackage{stfloats}

\algrenewcommand{\algorithmicrequire}{\textbf{Input:}}
\algrenewcommand{\algorithmicensure}{\textbf{Output:}}

\usepackage[colorlinks=true,linkcolor=blue,citecolor=blue,urlcolor=blue]{hyperref}
\usepackage[capitalize,noabbrev]{cleveref}
\usepackage{pifont}

\begin{document}

\title{RGSQ: Riemannian Geometry-Sensitive Quantization for Large Vision-Language Models}


\author{Zhiping Wu, Dongdong Ren, Yangchengyu Zhou, Zhengjie Zhang, Wenbin Li, Hongbing Pan, Yang Gao
\IEEEcompsocitemizethanks{
\IEEEcompsocthanksitem
Zhiping Wu and Hongbing Pan are with the School of Electronic Science and Engineering, Nanjing University, Nanjing 210093, China (e-mail: zhipingwu@smail.nju.edu.cn; phb@nju.edu.cn).
\IEEEcompsocthanksitem
Dongdong Ren is with Geely Automobile Research Institute (Ningbo) Co., Ltd., Ningbo 315000, China (e-mail: Dongdong.Ren2@geely.com)
\IEEEcompsocthanksitem
Yangchengyu Zhou, Zhengjie Zhang, Wenbin Li, and Yang Gao are with the State Key Laboratory for Novel Software Technology and the School of Intelligence Science and Technology, Nanjing University, Suzhou Campus, Suzhou 215163, China (e-mail: liwenbin@nju.edu.cn; gaoy@nju.edu.cn).
}
}

\markboth{Riemannian Geometry-Sensitive Quantization for Large VLMs,~Vol.~14, No.~8, July~2027}%
{Wu \MakeLowercase{\textit{et al.}}: A Sample Article Using IEEEtran.cls for IEEE Journals}


\maketitle
\begin{abstract}
Large vision-language models (VLMs) can be efficiently deployed under stringent memory and latency constraints through post training quantization (PTQ).
However, most PTQ methods are designed for unimodal large language models (LLMs). These methods treat quantization errors as isotropic perturbations under the Euclidean assumption, which provides weak guidance on directions most sensitive to quantization in VLMs.
Consequently, directly adapting unimodal PTQ approaches or solely employing modality-specific scaling often leads to uneven bit-width distribution and inconsistent performance in low-bit settings.
To address these challenges, we propose \textbf{R}iemannian \textbf{G}eometry-\textbf{S}ensitive \textbf{Q}uantization (\textbf{RGSQ}), which formulates quantization as a reconstruction problem under a unified Fisher–Riemannian metric. RGSQ identifies modality-specific sensitive directions via Riemannian manifold mappings built from modality-partitioned empirical Fisher factors and fused into a modality-aware Kronecker-structured metric. We then apply geometry-aligned rotations to reorient the local tangent frame, steering low-bit perturbations toward loss-insensitive axes. Finally, we apply a whitening transformation that maps the Riemannian objective to an equivalent Euclidean form, enabling standard unimodal PTQ methods to evaluate multimodal quantization error under their original assumptions.
Across an extensive and diverse set of mainstream VLM benchmarks, RGSQ achieves the highest accuracy and stability under extremely low-bit settings (W2A8 and W3A8). It outperforms VLM-aware baselines, such as MBQ and MQuant, by up to 5.9\% and surpasses single-modality improvements by up to 8.6\%.
\end{abstract}


\begin{figure}[t]
    \centering
    \includegraphics[width=1\linewidth]{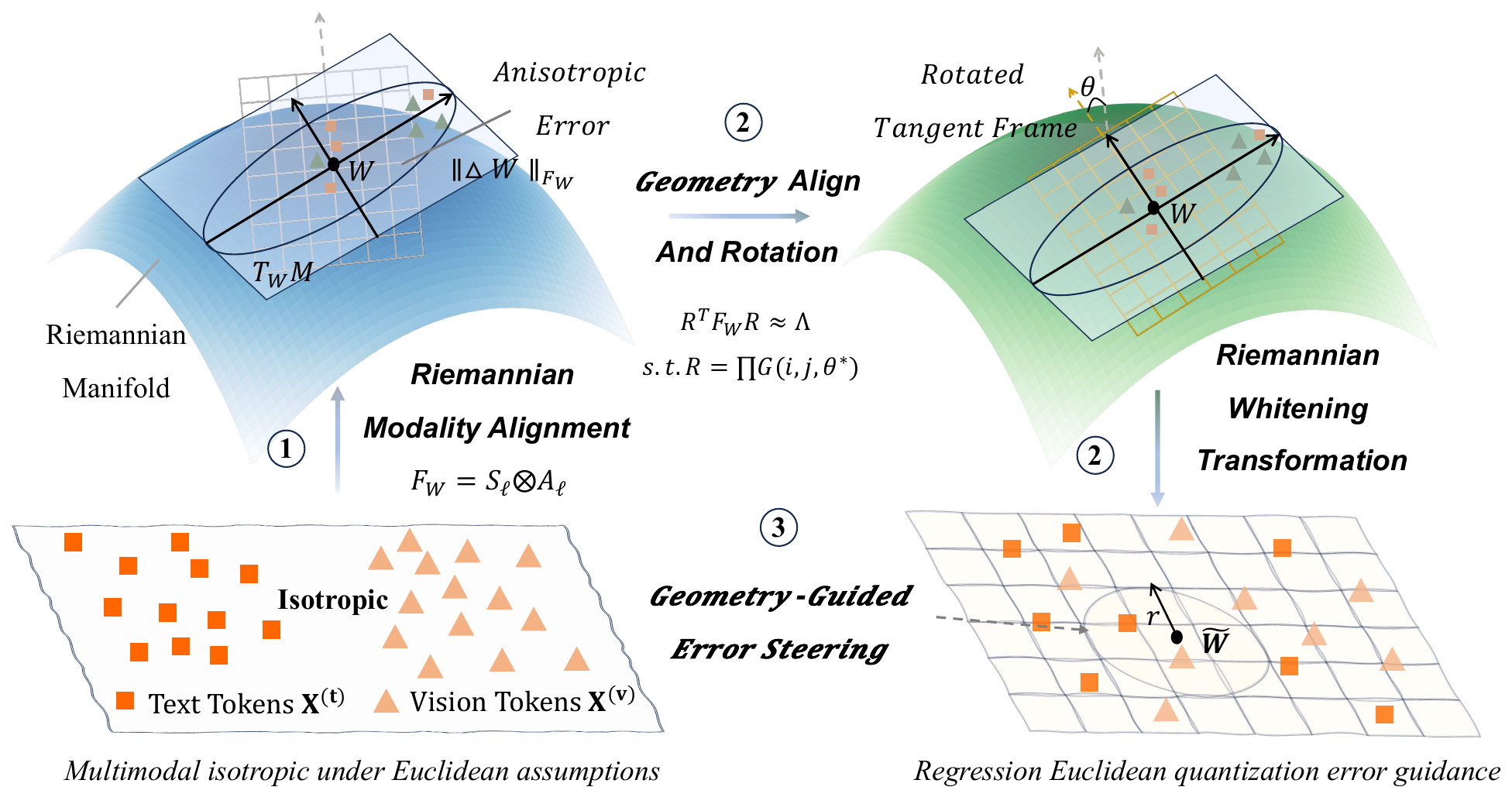}
    \caption{Illustration of the RGSQ framework: Rather than assuming multimodal quantization error is Euclidean and isotropic, we characterize each layer with a modality-aware Riemannian metric that reveals anisotropic, loss-sensitive directions. RGSQ then applies geometry-aligned rotations to reorient the local tangent frame, steering low-bit perturbations toward low-sensitivity axes. Finally, Riemannian whitening maps the metric back to an equivalent Euclidean form, enabling plug-and-play use of standard PTQ solvers for efficient error scoring and optimization.}
    \label{fig:framework}
\vspace{-10pt}
\end{figure}

\section{Introduction}
\IEEEPARstart{L}{arge} language models (LLMs) have made remarkable progress across a broad range of domains, including natural language processing~\cite{brown2020language}, code generation~\cite{guo2023longcoder}, complex reasoning~\cite{clark2018think}, and multimodal understanding~\cite{reid2024gemini, Zhang2026Multi}.
Recent multi-modality VLMs continue to expand the perception, reasoning, and generation capabilities of foundation models across visual and textual inputs~\cite{Li2026Mini}. 
This progress, however, has come at the price of a rapid growth in parameter count and memory footprint.
For example, the LLaMA-3 70B model requires approximately 140\,GB of memory under FP16 precision, which makes its deployment on resource-constrained devices challenging~\cite{llama3metaai}.
Consequently, designing techniques that substantially reduce inference-time resource consumption without retraining, or with minimal additional overhead, has become a critical prerequisite for the practical deployment of large vision–language models (VLMs).

\begin{figure*}[!t]
    \centering
    \includegraphics[width=0.90\linewidth]{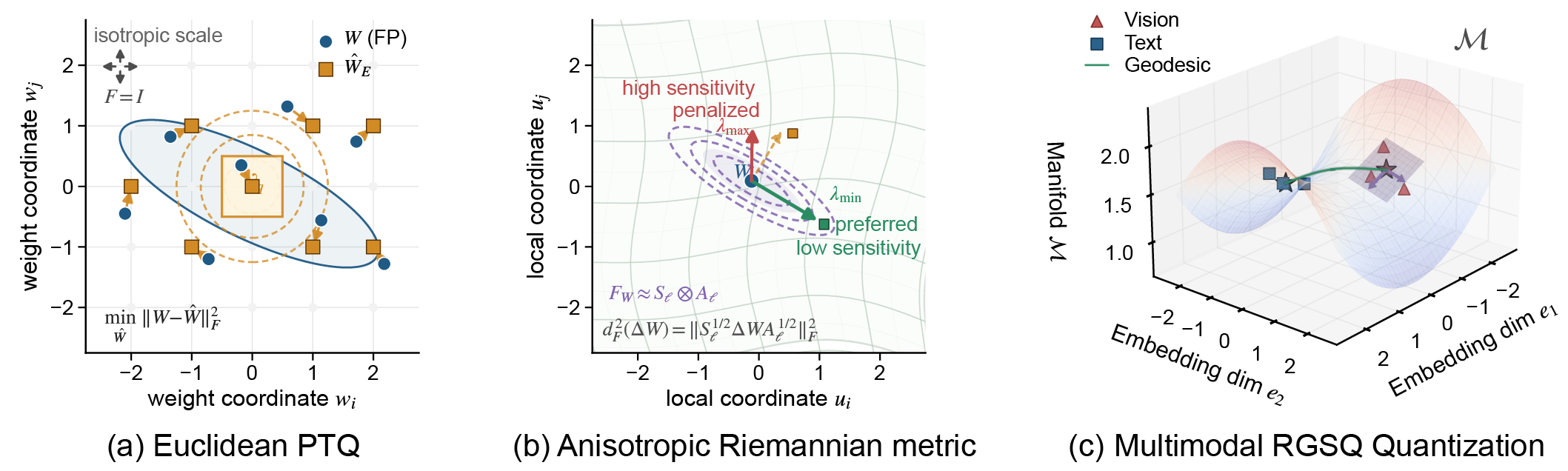}
    \caption{Illustration of the key motivation: (a) Euclidean PTQ quantifies distortion using an isotropic $\ell_{2}$ metric, which fails to capture anisotropic sensitivity. (b) A Riemannian metric makes this anisotropy explicit through curvature-aware distortion, but incurs substantially higher computational cost. (c) The proposed RGSQ reconciles this trade-off in VLMs by building a modality-aware unified Riemannian geometry that aligns cross-modal distortion scales with practical overhead, and by steering low-bit perturbations toward loss-insensitive directions.}
    \label{fig:motivation}
\vspace{-10pt}
\end{figure*}

Among efficiency-oriented techniques, including sparse attention~\cite{dao2022flashattention}, optimized decoding~\cite{cai2024medusa}, and quantization~\cite{dong2019hawq}, post-training quantization (PTQ) has been particularly attractive because it simultaneously reduces memory bandwidth and computational cost without requiring retraining.
Weight-only approaches such as AWQ~\cite{awq}, GPTQ~\cite{frantar2022gptq}, and QuIP~\cite{chee2024quip} primarily aim at relieving memory access and storage overhead, while weight–activation methods, including SmoothQuant~\cite{smoothquant}, SpinQuant~\cite{spinquant}, FlatQuant~\cite{flatquant}, and Atom further leverage low-precision tensor cores available on modern GPUs to accelerate inference.
To preserve downstream task performance, these methods typically rely on a calibration phase that optimizes scaling factors, channel equalization parameters, or rotation matrices by minimizing block-level reconstruction errors on a small calibration set.

Despite their success, the vast majority of existing PTQ methods are primarily designed for unimodal LLMs, and their direct application to vision–language models often leads to substantial accuracy degradation.
A few recent studies have begun to address this issue: Q-VLM~\cite{qvlm} estimates cross-layer dependencies via entropy and guides weighted quantization of visual encodings. MBQ~\cite{li2025mbq} introduces a modality-balanced strategy based on quantization sensitivity, and MQuant~\cite{yu2025mquant} combines modality-specific static quantization with attention-invariant flexible switching.
Although these methods mitigate cross-modal mismatches by correcting errors in Euclidean space, they overlook the quantization-sensitivity information encoded by natural gradients in high-dimensional parameter spaces.
A closely related effort, RSAVQ~\cite{xu2025rsavq}, introduces a Riemannian sensitivity-aware vector quantization formulation for LLMs, but it does not model the geometric relationships between modalities on the underlying manifold.
Table~\ref{tab:comparison} summarizes the conceptual differences between these representative methods and our work.

Our motivation is illustrated in Fig.~\ref{fig:motivation}.
We argue that the central challenge in quantizing VLMs originates from the simultaneous processing of visual and language features, which exhibit distinct statistical distributions and contribute unequally to the final task loss.
When the calibration process intermingles these heterogeneous modalities under a unified, isotropic optimization objective, limited calibration capacity is often allocated to less critical components.
This misallocation reduces representational fidelity for the most sensitive features and ultimately degrades overall model performance.

Based on this observation, we identify two fundamental challenges.
\emph{First}, VLMs exhibit cross-modal inconsistencies in sensitivity and second-order structure, so a single Euclidean reconstruction objective is inadequate for prioritizing different modalities.
Visual and language tokens differ not only in scale, but also in local curvature and loss sensitivity, assuming uniform token importance distorts the true cross-modal geometry and compromises the precision of highly sensitive components.
\emph{Second}, existing methods offer limited control over the directionality of low-bit quantization errors in multimodal settings.
In low-bit regimes, quantization errors are characterized not only by their magnitude but also by their direction, which critically affects the task loss.
Although RSAVQ accounts for directional errors, it is limited to unimodal scenarios and does not optimize cross-modal error directions, allowing quantization noise to accumulate and degrade performance in VLMs.

To address these challenges, we propose Riemannian Geometry-Sensitive Quantization (\textbf{RGSQ}) for VLMs, as illustrated in Fig.~\ref{fig:framework}.
RGSQ adopts an information-geometric perspective and constructs a modality-aware unified Riemannian metric for every linear layer.
Specifically, we estimate modality-specific second-order statistics, derive sensitivity-based aggregation weights from token-level gradient magnitudes, and fuse them into a unified Kronecker–Fisher metric that explicitly encodes cross-modal differences within a shared geometric space.
Under this unified geometry, the conventional Euclidean reconstruction objective is reformulated as Riemannian perturbation minimization, enabling principled handling of cross-modal sensitivity disparities.
Furthermore, to explicitly control quantization error directions, we introduce a geometry-guided error orientation mechanism that reparameterizes quantization errors via sparse Givens rotations.
This mechanism steers the unavoidable quantization noise toward directions of low sensitivity and low curvature, so that even though quantization errors remain, their impact on task loss is reduced—enabling stable and efficient low-bit quantization of VLMs.

VLMs are increasingly used as general-purpose tools across scientific document understanding, medical and biological image interpretation, and multimodal scientific question answering~\cite{liu2025septq}. However, their practical use in these high-stakes scenarios is constrained by both deployment costs and reliability risks, including tight compute, energy, and privacy budgets as well as relationship hallucinations and distraction-sensitive reasoning~\cite{Wu2026Evaluating,Yang2026Defying}. This shifts the central challenge from raw model capability alone to whether multimodal capability can be preserved under resource-limited and reliability-sensitive conditions. RGSQ is motivated by this requirement and aims to support stable low-bit deployment of VLMs.

Our main contributions can be summarized as follows.
\begin{itemize}
    \item We propose a unified Riemannian PTQ framework for VLMs that constructs a single, modality-aware Kronecker–Fisher metric from modality-decomposed second-order statistics and sensitivity-based weights, thereby resolving cross-modal imbalance in quantization error and ensuring equitable allocation of calibration capacity across modalities.

    \item We introduce a geometry-guided error orientation mechanism that uses sparse Givens rotations to reshape the basis in which quantization is performed. This enables joint control over both the magnitude and the direction of quantization noise, leading to substantially improved stability under low-bit settings.

    \item We validate RGSQ on three representative VLM families across model scales from 7B to 72B and on six mainstream vision–language benchmarks. RGSQ delivers a superior accuracy–efficiency trade-off, with particularly notable gains under low-bit and weight–activation joint quantization, while remaining hardware-friendly through the use of sparse rotations.
\end{itemize}

\section{Related Work}
\label{sec:related}

\subsection{Post-Training Quantization for LLMs}
\label{subsec:ptq_llm}
Post-training quantization (PTQ) has emerged as the dominant route to deploying large language models, because it substantially reduces inference-time memory footprint and bandwidth demand with little or no retraining overhead~\cite{Wang2026FinePruner}. In general, PTQ relies on a small set of quantization factors, such as scaling and zero-point offsets, to map high-precision floating-point tensors into low-bit integer formats, thereby compressing weights and intermediate activations and better exploiting the low-precision tensor cores available on modern hardware~\cite{Zhao2025LRQuant}. Existing PTQ methods for LLMs can be broadly grouped into scalar and vector quantization.

\textbf{Scalar quantization} learns per-channel or per-group scales and performs element-wise quantization for weights and activations, typically together with calibration-time compensation and outlier handling to reduce reconstruction error. Representative approaches include GPTQ~\cite{gptq} and AWQ~\cite{awq}, as well as distribution-shaping transformations such as rotations to improve quantization friendliness, e.g., QuaRot~\cite{quarot}, OSTQuant~\cite{hu2025ostquant}, and architecture-specific adaptations such as MambaQuant~\cite{xumambaquant}. Although these methods are simple and inference-friendly, they operate under Euclidean error surrogates during calibration, making it difficult to systematically capture direction-dependent sensitivity and structural heterogeneity that become pronounced at extremely low bit-widths.

\textbf{Vector quantization}, in contrast, encodes weight blocks or vectors using shared codebooks and indices, enabling higher compression ratios with improved representational capacity. Typical examples include GPTVQ~\cite{vangptvq}, VPTQ~\cite{liu2024vptq}, AQLM~\cite{egiazarian2024extreme}, CRVQ~\cite{xu2025crvq}, and QuIP\#~\cite{tseng2024quip}, which leverage codebook learning, residual quantization, or structured encoding to better fit quantization errors under aggressive compression. More recently, RSAVQ~\cite{xu2025rsavq} emphasizes sensitivity-aware and geometry-motivated error characterization, offering a finer-grained perspective for low-bit quantization.

However, LLM-centric PTQ techniques often degrade substantially when transferred directly to multimodal models. The underlying reason is that they implicitly assume relatively homogeneous distributions and representations, and therefore tend to apply shared scaling and unified calibration objectives across submodules. In VLMs, the visual and language branches differ markedly in their statistics, outlier patterns, and contributions to the final task loss. Treating multimodal weights and activations as a single modality—e.g., by sharing scaling factors or calibration targets—can amplify quantization errors on the more sensitive modality and lead to significant performance drops, which our experimental analysis (Section~\ref{sec:exp_main}) corroborates.

\subsection{Post-Training Quantization for VLMs}
\label{subsec:ptq_vlm}
To mitigate this modality mismatch, recent work has developed PTQ techniques specifically tailored for VLMs, where the central strategy is modality-aware scaling and calibration. Representative methods include Q-VLM~\cite{qvlm}, QSLAW~\cite{xie2024advancing}, MBQ~\cite{li2025mbq}, and MQuant~\cite{yu2025mquant}. These approaches either stabilize calibration through block- or partition-based quantization and cross-layer dependency modeling, or explicitly decouple visual and language components by assigning separate scaling factors and calibration weights. Many of them also exploit gradient- or loss-related signals as heuristic sensitivity indicators, reweighting reconstruction objectives to balance error across modalities~\cite{xiang2026fine}. Collectively, these studies demonstrate that explicitly accounting for modality discrepancies is necessary and can partially alleviate multimodal quantization collapse.

Nevertheless, most existing multimodal PTQ methods still rely on Euclidean error modeling and gradient-heuristic guidance: they minimize reconstruction losses defined in an implicit Euclidean space, and use back-propagated gradients to estimate importance for adjusting quantization strategies. The limitation is that Euclidean gradients are local and metric-dependent, and may not reliably reflect loss-consistent sensitive directions in high-dimensional parameter spaces—especially when multimodal interactions and low-bit error accumulation coexist. In multimodal settings, quantization priority is determined not only by token-scale differences, but more fundamentally by modality-induced local curvature and second-order sensitivity structures. Ignoring this can systematically misallocate limited calibration capacity across modalities.
As a result, purely Euclidean reconstruction objectives or gradient reweighting often fail to provide a consistent characterization of cross-modal second-order discrepancies and can be unstable under lower bit-widths and more challenging multimodal tasks.

\subsection{Geometry-Aware Quantization}
\label{subsec:geom_quant}
Information geometry and geometric deep learning suggest that the parameter spaces of neural networks exhibit pronounced non-Euclidean structures~\cite{dages2025finsler}. A Riemannian metric induced by the Fisher Information Matrix (FIM) provides a principled measure of how parameter perturbations affect a model's output distribution and task loss. In this view, natural gradients can be interpreted as scale-consistent updates along the underlying manifold, while approximations such as K-FAC~\cite{martens2015optimizing} make second-order geometric modeling tractable for large networks. Bringing this perspective to quantization implies that quantization errors should not be assessed solely by Euclidean distances, but rather by geometry-consistent perturbation costs, and that one should seek perturbation directions that are minimally harmful to the task loss.

Along this line, RSAVQ~\cite{xu2025rsavq} shows that geometry- and sensitivity-aware formulations can better explain low-bit performance degradation and enable direction-dependent error analysis. However, existing geometry-aware quantization studies are largely developed for unimodal settings or isolated structures, and do not explicitly resolve two key questions in multimodal quantization. \emph{First}, \textit{how can weights and activations from different modalities be mapped into a unified Riemannian geometric basis so that cross-modal second-order discrepancies and sensitivity inconsistencies can be modeled within a comparable metric space?} \emph{Second}, \textit{how can one exploit modality-specific sensitive directions in this geometric space to impose directional constraints on quantization noise, and then map the result back to a standard Euclidean computation graph to remain compatible with inference-friendly PTQ pipelines?} In the low-bit regime, both the directionality and the accumulation path of quantization errors substantially influence the final loss, and Euclidean reconstruction or gradient-heuristic reweighting alone are insufficient to stably control the error trajectory. A more systematic approach for multimodal low-bit deployment should therefore re-formulate PTQ objectives under a unified Riemannian metric and explicitly control error directions, steering unavoidable quantization noise toward low-sensitivity and low-curvature directions to obtain a more reliable accuracy efficiency trade-off.

This is precisely the gap addressed by RGSQ, which does not treat quantization error as an isotropic perturbation but instead measures it under a unified, modality-aware Fisher–Riemannian metric and steers it via sparse Givens rotations onto geometry-insensitive axes. As detailed in Section~\ref{sec:method}, this view subsumes both Euclidean PTQ and scalar modality-reweighting methods as special cases.

\section{Preliminaries}
\label{sec:prelim}
\algrenewcommand{\algorithmicrequire}{\textbf{Input:}}
\algrenewcommand{\algorithmicensure}{\textbf{Output:}}
\subsection{PTQ for LLMs and VLMs}
\label{sec:prelim-ptq}

\textbf{VLM architecture and token notation.}
A typical VLM consists of a visual encoder, a vision–language projector, and a large language model (LLM) backbone. The visual encoder processes input images or video frames into compact patch features. The projector aligns these visual features with the text embedding space, producing visual tokens $\mathbf{X}^{(v)}$. Text inputs are tokenized and embedded into text tokens $\mathbf{X}^{(t)}$. These multimodal tokens are then concatenated and fed into a decoder-style LLM that generates output sequences autoregressively.

At each Transformer layer $\ell$, we denote the token set as $\mathcal{T}_{\ell}$, which is naturally partitioned by modality:
$\mathcal{T}_{\ell} = \bigcup_{m\in\mathcal{M}} \mathcal{T}_{\ell}^{(m)}$,
where $m\in\mathcal{M}$ indexes the modalities (e.g., vision $v$ and text $t$).
This partition is inherent to VLMs, since visual tokens arise from the projector while text tokens arise from the embedding table.

We adopt symmetric uniform quantization for both weights and activations. For a scalar $z\in\mathbb{R}$, scale $s>0$, and bit-width $b$, the quantization operator is
\begin{equation}
    Q(z;s,b) = s\cdot \mathrm{clip}\!\left( \mathrm{round}\!\left(\tfrac{z}{s}\right), -2^{b-1},\, 2^{b-1}-1 \right),
\label{eq:uniform_quant}
\end{equation}
which is applied per channel or per group for tensors. We denote quantized weights and activations by $\widehat{W}=Q(W),\ \widehat{X}=Q(X)$, and the corresponding perturbations by $\Delta W=W-\widehat{W}$ and $\Delta X=X-\widehat{X}$.

\textbf{Euclidean PTQ objective.}
For a linear projection $Y_{\ell}=W_{\ell}X_{\ell}$, common Euclidean PTQ minimizes the output reconstruction error over a calibration set $\mathcal{D}_{\mathrm{cal}}$:
\begin{equation}
    \min_{\widehat{W}_{\ell}}
    \ \mathbb{E}_{(x,y)\in\mathcal{D}_{\mathrm{cal}}}
    \big[ \|W_{\ell}X_{\ell}-\widehat{W}_{\ell}X_{\ell}\|_{F}^{2} \big],
\label{eq:euclid_ptq}
\end{equation}
or one of its group-wise / channel-wise variants. This formulation implicitly assumes that Euclidean deviations are a faithful surrogate for the change in task loss and that all token modalities should be measured under a single isotropic scale. Both assumptions are systematically violated in VLMs.

\subsection{Information-Geometric View of Quantization Error}
\label{sec:prelim-geom}

\textbf{Riemannian metric from information geometry.}
A geometric view of quantization relates parameter perturbations to changes in the model's output distribution~\cite{song2025riemannian,park2026adaptive}.
Let $p_{\theta}(\cdot)$ denote the model distribution parameterized by $\theta$, and consider a small perturbation $\Delta\theta$.
A second-order Taylor expansion of the log-likelihood $\ell(\theta;u)=\log p_{\theta}(u)$ gives
\begin{equation}
    \ell(\theta\!+\!\Delta\theta;u) \approx \ell(\theta;u)
    + \nabla_{\theta}\ell(\theta;u)^{\!\top}\Delta\theta
    + \tfrac{1}{2}\Delta\theta^{\!\top}\!\nabla_{\theta}^{2}\ell(\theta;u)\Delta\theta.
\end{equation}
Under standard regularity conditions, $\mathbb{E}[\nabla_{\theta}\ell(\theta;u)]\!=\!0$, so the first-order term vanishes in expectation. Combining this with the Bartlett identity $F_{\theta} \!=\! -\mathbb{E}[\nabla_{\theta}^{2}\ell(\theta;u)]$ yields the local quadratic approximation
\begin{equation}
    D_{\mathrm{KL}}\big(p_{\theta}\,\|\,p_{\theta+\Delta\theta}\big) \approx \tfrac{1}{2}\Delta\theta^{\!\top}F_{\theta}\Delta\theta,
\label{eq:kl_quadratic}
\end{equation}
where $F_{\theta}$ is the Fisher Information Matrix (FIM)
\begin{equation}
    F_{\theta} = \mathbb{E}\big[ \nabla_{\theta}\log p_{\theta}(u)\ \nabla_{\theta}\log p_{\theta}(u)^{\!\top} \big].
\label{eq:fisher_def}
\end{equation}
Because $F_{\theta}$ is positive semi-definite, it serves as a canonical Riemannian metric tensor that measures distribution shifts in directions sensitive to local loss changes, inducing the inner product $\langle \Delta\theta_{1},\Delta\theta_{2}\rangle_{\theta} = \Delta\theta_{1}^{\!\top}F_{\theta}\Delta\theta_{2}$.

\textbf{Kronecker-factored curvature for linear layers.}
For a linear layer $Y\!=\!WX$ with loss $\mathcal{L}$, denote the output gradient by $G=\partial \mathcal{L}/\partial Y$. A widely used curvature proxy for $W$ is the \emph{empirical Fisher}
\begin{equation}
    F_{W} = \mathbb{E}\big[\mathrm{vec}(\nabla_{W}\mathcal{L})\,\mathrm{vec}(\nabla_{W}\mathcal{L})^{\!\top}\big].
\end{equation}
Using $\nabla_{W}\mathcal{L} = G X^{\!\top}$ and the identity $\mathrm{vec}(G X^{\!\top}) = (X\otimes I)\,\mathrm{vec}(G)$, a separable second-order structure follows once cross-covariances are ignored~\cite{martens2015optimizing}:
\begin{equation}
    F_{W} \approx S \otimes A,
\label{eq:kfac}
\end{equation}
where $A = \mathbb{E}_{x \sim \mathcal{D}_{\mathrm{cal}}} [xx^{\!\top}]$ and $S = \mathbb{E}_{g \sim \mathcal{D}_{\mathrm{cal}}} [gg^{\!\top}]$. The Kronecker approximation reduces the cost of storing and manipulating $F_{W}$ from $\mathcal{O}((d_{\mathrm{out}}d_{\mathrm{in}})^{2})$ to $\mathcal{O}(d_{\mathrm{out}}^{2}\!+\!d_{\mathrm{in}}^{2})$ while retaining anisotropic sensitivity from both inputs and gradients.

\textbf{Why the geometric view matters for VLMs.}
In VLMs, visual and language tokens often exhibit distinct activation ranges, outlier patterns, and gradient statistics. A single Euclidean norm flattens this anisotropy and treats heterogeneous modalities as if they shared a common scale. By contrast, a Fisher–Riemannian metric exposes modality-dependent sensitive directions through the curvature factors $A$ and $S$. This insight directly motivates the modality-aware geometric construction in Section~\ref{sec:metric}.

\begin{table}[t]\small
\centering
\caption{Comparison of quantization frameworks. 
(a) \textbf{Euclidean PTQ} minimizes isotropic quantization error, assuming uniform sensitivity ($I$) across all parameters. 
(b) \textbf{Standard Riemannian methods} utilize anisotropic curvature information ($S \otimes A$) but typically apply uniform weighting to all tokens, ignoring modality differences. 
(c) \textbf{RGSQ (Ours)} constructs a \textit{modality-aware unified metric} ($S_l \otimes A_l$) by aggregating modality-partitioned statistics with sensitivity-based weighting ($\pi^{(m)}$), and further incorporates geometry-guided error steering into the objective.}
\label{tab:comparison}
\renewcommand{\arraystretch}{1} 
\setlength{\tabcolsep}{1pt} 
\resizebox{\linewidth}{!}{
\begin{tabular}{l|c|c|c}
\textbf{Attribute} & \textbf{(a) Euclidean} & \textbf{(b) Riemannian} & \textbf{(c) RGSQ (Ours)} \\
\midrule
\textbf{Objective} & \multirow{1}{*}{$\|\Delta W\|_F^2$} & \multirow{1}{*}{$\|\Delta W\|_{S \otimes A}^2$} & \multirow{1}{*}{$\|\Delta W\|_{S_{\ell} \otimes A_{\ell}}^2 + \lambda \mathcal{L}_{\mathrm{geom}}$} \\
\midrule
\textbf{Sensitivity} & $I$ & $S \otimes A$ & $S_{\ell} \otimes A_{\ell}$ \\
\midrule
\textbf{Weighting} & ($1$) & ($1/N$) & {\footnotesize $\pi^{(m)} \propto \mathbb{E} \big[\frac{1}{|\mathcal{T}^{(m)}|} \sum_t \|g_t\|_1 \big]$} \\

\end{tabular}
}
\end{table}


\section{Method}
\label{sec:method}

We propose \textbf{RGSQ}, a unified Riemannian post-training quantization framework tailored to large VLMs. The core idea is to construct a unified Riemannian metric tensor on the shared token interface, which enables us to (i) quantitatively evaluate the true impact of quantization noise on the task loss and output distribution shifts, and (ii) directionally steer the noise toward low-curvature, low-sensitivity axes. 


\subsection{Riemannian Modality-Aware Metric}
\label{sec:metric}
We first map the visual and text tokens from their original Euclidean space onto a Riemannian manifold, where perturbations that were previously treated as isotropic reveal their anisotropic, modality-dependent structure. We construct a unified metric for each linear layer $Y_{\ell} = W_{\ell} X_{\ell}$ by estimating modality-wise second-order factors, weighting them by loss sensitivity, and fusing them into a unified Kronecker–Fisher form.

\textbf{Modality-partitioned second-order factors.}
Let $X_{\ell} \in \mathbb{R}^{d_{\mathrm{in}} \times N}$ and $Y_{\ell} \in \mathbb{R}^{d_{\mathrm{out}} \times N}$ denote stacked token activations and pre-activations on a calibration batch with $N$ tokens, and let $G_{\ell} = \partial \mathcal{L} / \partial Y_{\ell}$. For each modality $m$, we extract the submatrices $X_{\ell}^{(m)}$ and $G_{\ell}^{(m)}$ corresponding to tokens in $\mathcal{T}_{\ell}^{(m)}$, and compute the modality-partitioned empirical estimates:
\begin{equation}
\begin{aligned}
    A_{\ell}^{(m)} &= \tfrac{1}{N_m}\, X_{\ell}^{(m)} (X_{\ell}^{(m)})^{\!\top}, \\
    S_{\ell}^{(m)} &= \tfrac{1}{N_m}\, G_{\ell}^{(m)} (G_{\ell}^{(m)})^{\!\top},
\end{aligned}
\label{eq:modality_stats}
\end{equation}
where $N_m = |\mathcal{T}_{\ell}^{(m)}|$. The pair $(A_{\ell}^{(m)}, S_{\ell}^{(m)})$ provides modality-conditioned second-order statistics of activations and gradients, and instantiates a modality-aware curvature proxy at layer $\ell$.

\textbf{Sensitivity-aware modality weighting.}
A modality with larger loss gradients at a given layer is typically more sensitive to perturbations at that layer.
Let $g_{\ell,t}$ denote the token-level gradient vector for token $t$ at layer $\ell$. We define the modality sensitivity score as
\begin{equation}
    \alpha_{\ell}^{(m)} = \mathbb{E}\Big[ \tfrac{1}{|\mathcal{T}_{\ell}^{(m)}|} \sum_{t \in \mathcal{T}_{\ell}^{(m)}} \| g_{\ell,t} \|_{1} \Big],
\label{eq:alpha_def}
\end{equation}
which uses the $\ell_{1}$ norm to emphasize sparse but strong gradient coordinates—often more indicative of quantization-critical channels than $\ell_{2}$ energy. We then normalize these scores into convex weights
\begin{equation}
    \pi_{\ell}^{(m)} = \frac{\alpha_{\ell}^{(m)}}{\sum_{m'\in\mathcal{M}}\alpha_{\ell}^{(m')}+\varepsilon},
    \qquad \varepsilon>0,
\label{eq:pi_def}
\end{equation}
where the small $\varepsilon$ ensures numerical stability for near-zero gradients early in calibration. By construction, $\pi_{\ell}^{(m)}\ge 0$ and $\sum_{m}\pi_{\ell}^{(m)}\!\approx\! 1$.

Crucially, $\pi_{\ell}^{(m)}$ depends on per-token sensitivity rather than the raw token count, so the resulting metric is invariant by construction to the absolute visual-to-text token ratio. In practice we further apply exponential moving-average smoothing across calibration batches and clamp $\alpha_{\ell}^{(m)}$ to suppress outliers. When a modality has very few tokens but should not be neglected, a token-count–balanced variant $\pi_{\ell}^{(m)}\!\propto\!\alpha_{\ell}^{(m)}\cdot|\mathcal{T}_{\ell}^{(m)}|$ can be used.

\textbf{Unified Kronecker–Fisher metric.}
We aggregate the modality-specific factors into a single layer-wise metric:
\begin{equation}
\label{eq:unify_AS}
\begin{aligned}
    A_{\ell} &= \sum_{m \in \mathcal{M}} \pi_{\ell}^{(m)} A_{\ell}^{(m)}, \\
    S_{\ell} &= \sum_{m \in \mathcal{M}} \pi_{\ell}^{(m)} S_{\ell}^{(m)},
\end{aligned}
\end{equation}
and define the unified curvature proxy as $F_{W_{\ell}}\!\approx\! S_{\ell}\!\otimes\! A_{\ell}$.
Because $A_{\ell}$ and $S_{\ell}$ are convex combinations of positive semi-definite matrices, the unified metric $S_{\ell}\!\otimes\! A_{\ell}$ remains PSD, so the induced Riemannian norm is well defined.
The visual and textual modalities are thereby mapped into a shared Riemannian space: for any quantization perturbation $\Delta W_{\ell}$, the same metric tensor measures its Riemannian length regardless of modality, while modality-specific sensitivities are still encoded through $\pi_{\ell}^{(m)}$.
Algorithm~\ref{alg:rgsq_metric} constructs the unified Kronecker–Fisher metric (Eq.~\eqref{eq:unify_AS}) by online accumulation of modality-partitioned statistics, which is memory-efficient and requires only a single calibration pass with activation and gradient collection.
\begin{algorithm}[ht]
\caption{Modality-Aware Metric Construction}
\label{alg:rgsq_metric}
\small
\begin{algorithmic}[1]
\Require Calibration set $\mathcal{D}_{\mathrm{cal}}$, linear layers $\{W_\ell\}_{\ell=1}^{L}$
\Ensure Modality-aware metrics $\{A_\ell,S_\ell,\pi_\ell^{(m)}\}_{\ell=1}^{L}$

\For{$\ell=1$ to $L$}
    \State Initialize $A_\ell^{(m)},S_\ell^{(m)},\alpha_\ell^{(m)}$ for each modality $m$
    \For{each calibration batch}
        \State Run forward to obtain $X_\ell$ and $G_\ell$
        \State Split tokens into $\{\mathcal{T}_\ell^{(m)}\}_{m\in\mathcal{M}}$
        \State Estimate modality factors:
        \Statex \hspace{\algorithmicindent}
        $A_\ell^{(m)}\!\leftarrow\!\frac{1}{N_m}X_\ell^{(m)}(X_\ell^{(m)})^\top,\quad 
        S_\ell^{(m)}\!\leftarrow\!\frac{1}{N_m}G_\ell^{(m)}(G_\ell^{(m)})^\top$
        \State Estimate sensitivity:
        \Statex \hspace{\algorithmicindent}
        $\alpha_\ell^{(m)}\!\leftarrow\!
        \mathbb{E}\!\left[
        \frac{1}{|\mathcal{T}_\ell^{(m)}|}
        \sum_{t\in\mathcal{T}_\ell^{(m)}}\|g_{\ell,t}\|_1
        \right]$
    \EndFor
    \State Normalize modality weights:
    \Statex \hspace{\algorithmicindent}
    $\pi_\ell^{(m)}\!\leftarrow\!
    \alpha_\ell^{(m)}
    /\big(\sum_{m'}\alpha_\ell^{(m')}+\varepsilon\big)$
    \State Fuse the unified metric:
    \Statex \hspace{\algorithmicindent}
    $A_\ell\!\leftarrow\!\sum_m\pi_\ell^{(m)}A_\ell^{(m)},\quad
    S_\ell\!\leftarrow\!\sum_m\pi_\ell^{(m)}S_\ell^{(m)}$
\EndFor
\State \Return $\{A_\ell,S_\ell,\pi_\ell^{(m)}\}_{\ell=1}^{L}$
\end{algorithmic}
\end{algorithm}

\subsection{Riemannian Objectives and Whitening}
\label{sec:objective_whiten}
We next turn the unified metric $F_{W_{\ell}}$ into a calibration-time PTQ objective.
This objective minimizes quantization perturbations under the induced Riemannian norm, yet we deliberately leverage the Kronecker structure to derive an equivalent Euclidean reduction via whitening.
This reduction preserves the anisotropic, loss-sensitive weighting encoded in $(S_{\ell}, A_{\ell})$ while enabling seamless reuse of standard Euclidean PTQ solvers, and without adding runtime operators.

\textbf{Riemannian objective for weight quantization.}
Let $\widehat{W}_{\ell}=Q(W_{\ell})$ be the quantized weights and $\Delta W_{\ell}=W_{\ell}-\widehat{W}_{\ell}$ the weight error.
Under the unified curvature proxy, we minimize the Riemannian norm of $\Delta W_{\ell}$:
\begin{equation}
    \min_{\widehat{W}_{\ell}}
    \ \|\Delta W_{\ell}\|_{F_{W_{\ell}}}^{2} = \min_{\widehat{W}_{\ell}}
    \ \mathrm{vec}(\Delta W_{\ell})^{\!\top}(S_{\ell}\!\otimes\! A_{\ell})\,\mathrm{vec}(\Delta W_{\ell}).
\label{eq:riem_weight_obj}
\end{equation}
Exploiting the Kronecker factorization, this quadratic form admits an equivalent Frobenius expression in whitened coordinates:
\begin{equation}
    \|\Delta W_{\ell}\|_{S_{\ell}\otimes A_{\ell}}^{2} = \|S_{\ell}^{1/2}\,\Delta W_{\ell}\,A_{\ell}^{1/2}\|_{F}^{2}.
\label{eq:riem_frob}
\end{equation}
A short derivation is instructive. Using the matrix identity $\mathrm{vec}(UVW)=(W^{\!\top}\!\otimes\! U)\,\mathrm{vec}(V)$, we have
\begin{equation}\small
\mathrm{vec}\!\left(S_{\ell}^{1/2}\Delta W_{\ell}A_{\ell}^{1/2}\right)
= \left((A_{\ell}^{1/2})^{\!\top}\!\otimes\! S_{\ell}^{1/2}\right)\mathrm{vec}(\Delta W_{\ell}),
\end{equation}
and therefore
\begin{equation}\small
\begin{aligned}
&\big\|S_{\ell}^{1/2}\Delta W_{\ell}A_{\ell}^{1/2}\big\|_{F}^{2}\\
&=
\mathrm{vec}(\Delta W_{\ell})^{\!\top}
\!\left(A_{\ell}^{1/2}\!\otimes\! S_{\ell}^{1/2}\right)^{\!\top}\!\!
\left(A_{\ell}^{1/2}\!\otimes\! S_{\ell}^{1/2}\right)\!
\mathrm{vec}(\Delta W_{\ell}) \nonumber\\
&=
\mathrm{vec}(\Delta W_{\ell})^{\!\top}(S_{\ell}\!\otimes\! A_{\ell})\,\mathrm{vec}(\Delta W_{\ell}),
\end{aligned}
\end{equation}
which recovers Eq.~\eqref{eq:riem_weight_obj}.
Because the empirical covariances $S_{\ell}$ and $A_{\ell}$ may be rank-deficient when the calibration set is small, in practice we add a small diagonal damping term, $\tilde{S}_{\ell} = S_{\ell} + \delta I$ and $\tilde{A}_{\ell} = A_{\ell} + \delta I$, before taking matrix square roots. This guarantees invertibility and numerical stability without altering the qualitative geometry.

\textbf{Whitening and Euclidean reduction.}
Define the geometry-aligned variables
$\widetilde{W}_{\ell}=S_{\ell}^{1/2}W_{\ell}A_{\ell}^{1/2}$ and
$\widetilde{\widehat{W}}_{\ell}=S_{\ell}^{1/2}\widehat{W}_{\ell}A_{\ell}^{1/2}$.
Substituting into Eq.~\eqref{eq:riem_frob} immediately gives
\begin{equation}
    \min_{\widehat{W}_{\ell}}
    \ \|S_{\ell}^{1/2}(W_{\ell}-\widehat{W}_{\ell})A_{\ell}^{1/2}\|_{F}^{2}
    \ \equiv\
    \min_{\widetilde{\widehat{W}}_{\ell}}
    \ \|\widetilde{W}_{\ell}-\widetilde{\widehat{W}}_{\ell}\|_{F}^{2}.
\label{eq:euclid_reduced}
\end{equation}
Eq.~\eqref{eq:euclid_reduced} reveals the central benefit of our formulation: any Euclidean PTQ solver that minimizes Frobenius reconstruction—such as GPTQ, AWQ, or SmoothQuant—can be applied directly to $\widetilde{W}_{\ell}$ and still optimize the original Riemannian objective. This yields the plug-and-play property exploited in our experiments (Section~\ref{sec:exp_row}).
Crucially, the whitening transformation is used only for calibration-time scoring and optimization. The deployed quantized model does not require additional runtime operators, since the whitening factors $S_{\ell}^{1/2}$ and $A_{\ell}^{1/2}$ are absorbed into the quantized parameters offline.

\textbf{Geometry-aware activation quantization.}
Activation quantization in VLMs is fragile because of modality-dependent ranges and sensitivities. Let $x_{\ell,t}$ denote the token activation vector at layer $\ell$ and token $t$. We calibrate scales by minimizing a modality-weighted reconstruction objective:
\begin{equation}
    \min_{s_{\ell}}
    \ \mathbb{E}_{t\in\mathcal{T}_{\ell}}
    \big[ \pi_{\ell}^{(m(t))} \|x_{\ell,t}-Q(x_{\ell,t};s_{\ell},b)\|_{2}^{2} \big],
\label{eq:act_obj}
\end{equation}
where $m(t)$ is the modality of token $t$. The weight $\pi_{\ell}^{(m)}$ allocates calibration capacity proportionally to modality sensitivity rather than to raw token frequency, which empirically improves stability over a single unweighted Euclidean scale.

\textbf{Global geometry regularization.}
Local reconstruction losses for weights and activations do not constrain the global semantic structure of the embedding space. Under aggressive low-bit quantization, even small per-layer errors can compound and distort representation similarities, harming downstream tasks that depend on cross-modal alignment. We therefore optionally regularize calibration so that the pairwise similarity structure of the calibration batch $\mathcal{B}$ is preserved. Let $\mathbf{z}_k\in\mathbb{R}^{d}$ be the representative embedding (e.g., pooled vision-language feature) for the $k$-th sample, and define the cosine similarities
\begin{equation}
s_{ij} = \frac{\mathbf{z}_i^{\!\top}\mathbf{z}_j}{\|\mathbf{z}_i\|_{2}\|\mathbf{z}_j\|_{2}},
\quad
\hat{s}_{ij} = \frac{\hat{\mathbf{z}}_i^{\!\top}\hat{\mathbf{z}}_j}{\|\hat{\mathbf{z}}_i\|_{2}\|\hat{\mathbf{z}}_j\|_{2}},
\end{equation}
for the full-precision and quantized models, respectively. To prevent collapse of the semantic geometry, we minimize the discrepancy between similarity matrices:
\begin{equation}
\mathcal{L}_{\mathrm{geom}}
= \frac{1}{|\mathcal{B}|^{2}}\sum_{i\in\mathcal{B}}\sum_{j\in\mathcal{B}}(s_{ij}-\hat{s}_{ij})^{2}.
\label{eq:global_geom}
\end{equation}
This pairwise constraint stabilizes calibration against local minima that yield small per-layer errors but large global distortions.

\subsection{Geometry-Guided Error Steering}
\label{sec:error_steering}
Whitening aligns the metric used to measure quantization error, yet it does not by itself ensure that the realized perturbations avoid the high-curvature, loss-sensitive directions of the unified metric. To directly shape the directionality of noise, we introduce error steering via lightweight sparse rotations that reparameterize the channel coordinates.

For each layer $\ell$, we apply an orthogonal transform $R_{\ell}$ on the input-channel dimension, parameterized as a product of sparse Givens layers:
\begin{equation}
    R_{\ell}=R_{\ell,K}\cdots R_{\ell,2}R_{\ell,1},
\label{eq:givens_prod}
\end{equation}
where each $R_{\ell,k}$ is block-diagonal and consists of disjoint $2\!\times\!2$ Givens rotations on non-overlapping channel pairs. Concretely, in the $k$-th layer we partition the channel indices into $P\!\le\!\lfloor d_{\mathrm{in}}/2\rfloor$ disjoint pairs $\{(u_{p},v_{p})\}_{p=1}^{P}$ and apply a planar rotation on each pair:
\begin{equation}
G(u,v;\theta)=
\begin{bmatrix}
\cos\theta & -\sin\theta\\
\sin\theta & \cos\theta
\end{bmatrix},
\end{equation}
while leaving all other channels unchanged. Thus $R_{\ell,k}$ contains $2\!\times\!2$ rotation blocks on the paired coordinates and identity blocks elsewhere, and $R_{\ell}$ is formed by composing $K$ such sparse layers.

For the linear projection $Y_{\ell}=W_{\ell}X_{\ell}$, we apply the orthogonal reparameterization
\begin{equation}
    X_{\ell}'=R_{\ell}^{\!\top}X_{\ell}, \qquad W_{\ell}'=W_{\ell}R_{\ell},
\label{eq:rot_reparam}
\end{equation}
which preserves the full-precision mapping since $W_{\ell}X_{\ell}\!=\!W_{\ell}'X_{\ell}'$.
We then quantize $W_{\ell}'$ and $X_{\ell}'$. By changing the basis in which quantization is performed, the induced low-bit perturbations become concentrated on geometry-insensitive directions under the unified metric, which we will show improves stability in the low-bit regime.

\textbf{Geometry-aware channel importance.}
Channel pairing for the sparse Givens layers must balance two factors: modality-dependent loss sensitivity and channel-wise curvature.
Let $h_{\ell,u}^{(m)}$ denote the diagonal curvature proxy of channel $u$ under modality $m$, estimated from the diagonal entries of the modality-partitioned Fisher factors $S_{\ell}^{(m)}$ and $A_{\ell}^{(m)}$. We then define
\begin{equation}
\label{eq:importance_score}
\begin{aligned}
\omega_{\ell,u}^{(m)}&=\pi_{\ell}^{(m)}\cdot h_{\ell,u}^{(m)}, \\
\tilde{\omega}_{\ell,u}&=\mathrm{Normalize}\Big(\textstyle\sum_{m}\omega_{\ell,u}^{(m)}\Big),
\end{aligned}
\end{equation}
which serves as a unified importance score that modulates a channel-wise curvature signal with the modality sensitivity weights $\pi_{\ell}^{(m)}$. The same scores also guide layer-wise quantization hyperparameters (group-wise clipping and scale calibration), so that directions with larger $\tilde{\omega}$ incur smaller quantization perturbations.

\textbf{Rotation selection objective.}
We select $R_{\ell}$ to minimize the unified quantization error under the Riemannian geometry in the reparameterized coordinates.
Let $\Delta W_{\ell}'=W_{\ell}'-\widehat{W}_{\ell}'$ denote the rotated weight error.
Because $W_{\ell}'=W_{\ell}R_{\ell}$, the input-side curvature factor updates as $A'_{\ell} = R_{\ell}^{\!\top} A_{\ell} R_{\ell}$.
We score a candidate $R_{\ell}$ using the whitening-based Euclidean reduction:
\begin{equation}
\label{eq:rotation_obj}
\begin{aligned}
    \mathcal{J}_{\ell}(R_{\ell})
    & = \|S_{\ell}^{1/2}\,\Delta W_{\ell}'\,(A'_{\ell})^{1/2}\|_{F}^{2} \\
    & \quad + \lambda_{A}\,
    \mathbb{E}_{t}\!\Big[ \pi_{\ell}^{(m(t))}\| x'_{\ell,t}-Q(x'_{\ell,t};s_{\ell},8) \|_{2}^{2} \Big],
\end{aligned}
\end{equation}
where $x'_{\ell,t}$ denotes the rotated activation at layer $\ell$. Because $R_{\ell}$ is sparse, computing $A'_{\ell}$ is efficient.

\textbf{Greedy discrete search.}
The joint optimization over all Givens pairs is combinatorial, so we adopt a greedy procedure.
For each layer $k\!\in\!\{1,\dots,K\}$, we iterate through the disjoint channel pairs $\mathcal{P}_{k}$ ordered by $\tilde{\omega}$, and for each pair $(i,j)$ perform a discrete grid search over a candidate angle set $\Theta$. In practice we use a uniform grid $\Theta = \{0,\,\pi/16,\,\dots,\,\pi\}$ to balance search precision and calibration time, and the selection criterion is the minimization of $\mathcal{J}_{\ell}$ defined in Eq.~\eqref{eq:rotation_obj}.
Algorithm~\ref{alg:rgsq_rotation} executes geometry-guided error steering, where sparse rotations are iteratively optimized to minimize the whitened quantization error (Eq.~\eqref{eq:rotation_obj}) under the sensitivity-weighted curvature importance score $\tilde{\omega}_{\ell,u}$ derived in Algorithm~\ref{alg:rgsq_metric}.
\begin{algorithm}[ht]
\caption{Geometry-Guided Sparse Rotation}
\label{alg:rgsq_rotation}
\small
\begin{algorithmic}[1]
\Require Weights $\{W_\ell\}$, metrics $\{A_\ell^{(m)},S_\ell^{(m)},\pi_\ell^{(m)}\}$, Givens depth $K$, angle set $\Theta$
\Ensure Sparse rotations $\{R_\ell\}_{\ell=1}^{L}$

\For{$\ell=1$ to $L$}
\State Initialize $R_\ell\leftarrow I$
    \State Compute input-channel importance:
    \Statex \hspace{\algorithmicindent}
    $h_{\ell,u}^{(m)}
    \leftarrow
    A_{\ell,uu}^{(m)}
    \cdot
    \mathrm{tr}(S_\ell^{(m)})/d_{\mathrm{out}}$
    \Statex \hspace{\algorithmicindent}
    $\tilde{\omega}_{\ell,u}
    \leftarrow
    \mathrm{Normalize}\!\left(
    \sum_m \pi_\ell^{(m)} h_{\ell,u}^{(m)}
    \right)$
    \For{$k=1$ to $K$}
        \For{each $(i,j)\in\mathcal{P}_k$}
        \State Select the angle by greedy search:
         \State   $\theta^\star
            \leftarrow
            \mathop{\arg\min}\limits_{\theta\in\Theta}
            \mathcal{J}_\ell\!\left(R_\ell G(i,j,\theta)\right)$
            \State Update $R_\ell\leftarrow R_\ell G(i,j,\theta^\star)$
        \EndFor
    \EndFor
\EndFor
\State \Return $\{R_\ell\}_{\ell=1}^{L}$
\end{algorithmic}
\end{algorithm}

\textbf{Complexity and inference overhead.}
A key practical advantage of the sparse Givens parameterization over dense rotation matrices is efficiency during calibration.
For curvature updates, updating a Hessian approximation with a dense rotation requires $\mathcal{O}(d^{3})$ operations, whereas a single Givens rotation only affects two rows and two columns, reducing the update cost to $\mathcal{O}(d)$. Across $K$ layers of Givens rotations covering $d/2$ pairs each, the total complexity is $\mathcal{O}(K d^{2})$, substantially lower than dense operations when $K\!\ll\! d$.
For metric evaluation, the whitening transformation $S_{\ell}^{1/2} \Delta W_{\ell}' (A'_{\ell})^{1/2}$ exploits the sparsity of $R_{\ell}$ via incremental updates of $A'_{\ell}$, avoiding repeated expensive matrix decompositions.
For inference, the weight rotation $W_{\ell}'\!=\!W_{\ell}R_{\ell}$ is pre-computed offline so that the deployed model stores $\widehat{W}_{\ell}'$ directly, with \textbf{zero} runtime overhead for weights. The activation rotation $X_{\ell}'\!=\!R_{\ell}^{\!\top}X_{\ell}$ must be applied online, but each Givens layer adds only $4K$ multiplications per channel because every block is a $2\!\times\!2$ matrix. This is computationally much cheaper than the dense matrix multiplications used by QuIP\# or SpinQuant ($d$ multiplications per channel), making RGSQ particularly hardware-friendly for latency-critical deployments.

\subsection{Overall Calibration Loss}
\label{sec:overall_calib}
We combine the three pieces—geometry-aware weight reconstruction, modality-weighted activation reconstruction, and optional global semantic regularization—into a single calibration objective:
\begin{equation}
\label{eq:final_cal}
\begin{aligned}
    \mathcal{L}_{\mathrm{cal}}
    & = \lambda_{W}\sum_{\ell}\|S_{\ell}^{1/2}(W_{\ell}'-\widehat{W}_{\ell}')\ (A'_{\ell})^{1/2}\|_{F}^{2} \\
    & \quad + \lambda_{A}\sum_{\ell}\mathbb{E}_{t}\big[\pi_{\ell}^{(m(t))}\|\Delta x'_{\ell,t}\|_{2}^{2}\big]
    + \lambda_{\mathrm{geom}}\mathcal{L}_{\mathrm{geom}},
\end{aligned}
\end{equation}
where the first term measures weight quantization perturbations under the unified Riemannian geometry in reparameterized coordinates, the second term calibrates activation quantization with modality sensitivity weights, and $\mathcal{L}_{\mathrm{geom}}$ is an optional regularizer used to stabilize calibration-time hyperparameter selection.
Algorithm~\ref{alg:rgsq_quant} performs quantization in the geometry-aligned space: by applying the whitening transformation defined by the updated curvature $A_{\ell}'$ (Eq.~\eqref{eq:euclid_reduced}), the Riemannian objective reduces to a tractable Euclidean form, so that low-bit perturbations align with the cross-modal sensitivity structure.
\begin{algorithm}[ht]
\caption{RGSQ Quantization}
\label{alg:rgsq_quant}
\small
\begin{algorithmic}[1]
\Require Weights $\{W_\ell\}$, metrics $\{A_\ell,S_\ell,\pi_\ell^{(m)}\}$, rotations $\{R_\ell\}$, bit-widths $b_W,b_A$
\Ensure Quantized weights $\{\widehat{W}'_\ell\}$ and activation scales $\{s_\ell\}$

\For{$\ell=1$ to $L$}
    \State Rotate weights and activations:
    \Statex \hspace{\algorithmicindent}
    $W'_\ell \leftarrow W_\ell R_\ell,\quad
    X'_\ell \leftarrow R_\ell^\top X_\ell$
    
    \State Rotate input metric:
    \Statex \hspace{\algorithmicindent}
    $A'_\ell \leftarrow R_\ell^\top A_\ell R_\ell$
    
    \State Quantize weights by the whitened PTQ objective:
    \Statex \hspace{\algorithmicindent}
    $\widehat{W}'_\ell
    \leftarrow
    \mathop{\arg\min}\limits_{\bar{W}\in\mathcal{Q}_{b_W}(W'_\ell)}
    \mathcal{E}^{W}_\ell(\bar{W};S_\ell,A'_\ell)$
    
    \State Calibrate activation scale by modality-weighted error:
    \Statex \hspace{\algorithmicindent}
    $s_\ell
    \leftarrow
    \mathop{\arg\min}\limits_{s}
    \mathcal{E}^{A}_\ell(s;\pi_\ell^{(m)},X'_\ell,b_A)$
\EndFor

\State \Return $\{\widehat{W}'_\ell\},\{R_\ell\},\{s_\ell\}$
\end{algorithmic}
\end{algorithm}

\textbf{Why RGSQ is more than a reweighting.}
Compared with scalar reweighting strategies such as MBQ, RGSQ does not merely rebalance a per-modality scalar in a Euclidean loss. Instead, it computes a full $d\!\times\!d$ covariance matrix $A_{\ell}^{(m)}$ per modality, capturing the complete directional structure of how each modality's activations are distributed. The unified metric $A_{\ell}=\sum_{m}\pi_{\ell}^{(m)}A_{\ell}^{(m)}$ is therefore itself a $d\!\times\!d$ matrix, not a scalar, and encodes which specific weight directions are sensitive to which modality. The Riemannian norm $\|S_{\ell}^{1/2}\Delta W A_{\ell}^{1/2}\|_{F}^{2}$ then penalizes perturbations proportionally to their actual impact on the loss landscape, rather than to their bare Euclidean magnitude. 

\section{Experiments}
\label{sec:exp}

\subsection{Experimental Setups}
\label{sec:exp_setups}
\begin{table*}[h]
\centering
\caption{Main results on the small models of LLaVA-onevision, InternVL2, and Qwen2-VL families. $\dagger$ denotes our reported results.}
  {\renewcommand{\arraystretch}{0.95}
  \resizebox{0.9\textwidth}{!}{
      \begin{tabular}{cccccccccc}
        \toprule
        Model & Bitwidth & Method & MMMU & SEED & OCRBench & VizWiz & ScienceQA & TextVQA & Average ($\uparrow$) \\
        \midrule
        \multirow{11}{*}{LLaVA-onevision-7B} & FP16 & - & 46.0 & 74.9 & 62.2 & 60.4 & 85.4 & 76.1 & 67.5\\
        \cmidrule(lr){2-10}
        ~ & \multirow{5}{*}{W3A16} & RTN & 34.7 & 5.9 & 35.9 & 59.2 & 86.2 & 60.9 & 47.1\\
        ~ & ~ & GPTQ & 41.9 & \textbf{72.9} & 55.7 & 56.4 & 86.4 & 71.3 & 64.1\\
        ~ & ~ & AWQ & 36.6 & 53.0 & 59.3 & 58.5 & 83.2 & 73.0 & 60.6\\
        ~ & ~ & MBQ & 42.0 & 69.7 & 61.1 & 60.7 & 85.0 & 73.3 & 65.3\\
        ~ & ~ & RGSQ & \textbf{47.5} & 72.0 & \textbf{61.8} & \textbf{60.2} & \textbf{88.5} & \textbf{75.2} & \textbf{67.5}\\
        \cmidrule(lr){2-10}
        ~ & \multirow{5}{*}{W4A8} & RTN & 38.2 & 50.3 & 40.1 & 58.2 & 88.3 & 61.5 & 56.1\\
        ~ & ~ & SQ & 30.9 & 42.7 & 32.0 & 56.7 & 87.1 & 56.9 & 51.1\\
        ~ & ~ & MBQ & 42.6 & 67.7 & 52.3 & 58.9 & 88.5 & 68.3 & 63.1\\
        ~ & ~ & MQuant$^\dagger$ & 45.3 & \textbf{73.7} & 58.8 & 59.4 & 84.0 & 74.9 & 66.0\\
        ~ & ~ & RGSQ & \textbf{48.4} & 72.4 & \textbf{62.0} & \textbf{60.1} & \textbf{89.9} & \textbf{75.9} & \textbf{68.1}\\
        \midrule
        \multirow{11}{*}{InternVL2-8B} & FP16 & - & 48.0 & 76.0 & 76.5 & 61.1 & 96.2 & 77.0 & 72.5\\
        \cmidrule(lr){2-10}
        ~ & \multirow{5}{*}{W3A16} & RTN & 43.7 & 74.9 & 74.0 & 56.0 & 95.6 & 74.6 & 69.8\\
        ~ & ~ & GPTQ & 41.7 & 73.4 & 70.2 & 59.9 & 89.5 & 73.1 & 68.0\\
        ~ & ~ & AWQ & 44.8 & 75.2 & 74.7 & 58.9 & 95.5 & 74.2 & 70.6\\
        ~ & ~ & MBQ & 46.9 & 75.4 & 75.1 & 58.7 & 95.6 & 75.1 & 71.1\\
        ~ & ~ & RGSQ & \textbf{47.6} & \textbf{75.8} & \textbf{76.0} & \textbf{60.5} & \textbf{96.0} & \textbf{76.5} & \textbf{72.1}\\
        \cmidrule(lr){2-10}
        ~ & \multirow{5}{*}{W4A8} & RTN & 44.3 & 74.0 & 72.0 & 57.1 & 95.5 & 73.1 & 69.3\\
        ~ & ~ & SQ & 41.8 & 73.8 & 70.9 & 54.2 & 95.1 & 72.6 & 68.1\\
        ~ & ~ & MBQ & 45.6 & 74.3 & 73.0 & 56.5 & 95.8 & 72.3 & 69.6\\
        ~ & ~ & MQuant$^\dagger$ & 46.9 & 74.3 & 69.9 & 59.8 & 94.1 & 76.0 & 70.2\\
        ~ & ~ & RGSQ & \textbf{48.2} & \textbf{76.0} & \textbf{76.4} & \textbf{61.0} & \textbf{96.4} & \textbf{77.0} & \textbf{72.5}\\
        \midrule
        \multirow{11}{*}{Qwen2-VL-7B} & FP16 & - & 50.6 & 76.4 & 80.7 & 68.3 & 85.1 & 82.0 & 73.8\\
        \cmidrule(lr){2-10}
        ~ & \multirow{5}{*}{W3A16} & RTN & 44.9 & 74.8 & 60.0 & 65.2 & 81.5 & 71.2 & 66.3\\
        ~ & ~ & GPTQ & 43.1 & 73.7 & 74.8 & 64.3 & 79.7 & 76.7 & 68.7\\
        ~ & ~ & AWQ & 44.7 & 75.1 & 76.9 & \textbf{68.0} & 82.5 & 79.5 & 71.1\\
        ~ & ~ & MBQ & 47.9 & 74.8 & 76.8 & 67.7 & 82.8 & 79.9 & 71.6\\
        ~ & ~ & RGSQ & \textbf{50.0} & \textbf{76.0} & \textbf{79.8} & 67.8 & \textbf{84.5} & \textbf{81.2} & \textbf{73.2}\\
        \cmidrule(lr){2-10}
        ~ & \multirow{5}{*}{W4A8} & RTN & 43.8 & 74.9 & 60.3 & 58.9 & 78.9 & 71.0 & 64.6\\
        ~ & ~ & SQ & 45.9 & 75.0 & 57.1 & 52.3 & 80.9 & 68.2 & 63.2\\
        ~ & ~ & MBQ & 47.2 & 75.1 & 72.8 & 59.3 & 81.2 & 75.0 & 68.4\\
        ~ & ~ & MQuant$^\dagger$ & 48.6 & 71.1 & 79.2 & 65.4 & 83.5 & 79.6 & 71.3\\
        ~ & ~ & RGSQ & \textbf{50.8} & \textbf{76.4} & \textbf{80.5} & \textbf{68.2} & \textbf{85.2} & \textbf{82.0} & \textbf{73.8}\\
        \bottomrule
      \end{tabular}
    }}
  \label{tab:main-small}
\end{table*}

\textbf{Calibration dataset.}
The calibration set leverages a VLM's joint perception–generation ability, which is naturally exercised by image-captioning data.
Specifically, we adopt the refined COCO Caption collection released as part of ShareGPT4V~\cite{sharegpt4v}, consistent with AWQ, MBQ, and MQuant.
Each calibration sample consists of (i) a single COCO image and (ii) a detailed caption generated by GPT-4V, formatted as a multi-turn conversation using the model-specific chat template.
During preprocessing, the image is encoded by the model's vision tower into visual tokens, while the caption is tokenized into roughly $64$–$256$ text tokens depending on length.
The two sequences are then concatenated into one input, ensuring that both modalities are present in every calibration forward pass.

\textbf{Evaluation benchmarks.}
We evaluate quantization quality on a broad suite of vision-language benchmarks through the LMMs-Eval framework~\cite{lmmseval}: OCRBench~\cite{ocrbench} and TextVQA~\cite{textvqa} for text recognition and understanding. VizWiz~\cite{vizwiz} and SEED-Bench~\cite{seed-bench} for general visual perception, and ScienceQA~\cite{scienceqa} together with MMMU~\cite{mmmu} for multimodal reasoning. To probe generalization beyond static images, we additionally evaluate on the SEED-Video~\cite{seed-bench} and Video-MME~\cite{fu2025video} splits.

\textbf{Models.}
We evaluate three representative VLM families: LLaVA-OneVision~\cite{llava-onevision}, InternVL2~\cite{internvl}, and Qwen2-VL~\cite{qwen-vl}.
Within each family we include one compact and one large-scale variant to verify scaling.
For LLaVA-OneVision we test the 7B and 72B variants, which pair Qwen2-7B/-72B as the language backbone with SigLIP-400M~\cite{siglip} as the ViT encoder.
For InternVL2 we consider the 8B and 26B models, built on InternLM2-8B/-20B with InternViT-300M/-6B as the visual encoder.
For Qwen2-VL we cover the 7B and 72B models, using Qwen2-7B/-72B with a 675M ViT encoder.

\textbf{Baselines.}
\begin{table}[t]\small
\centering
\caption{Default quantization hyperparameters used in all experiments unless otherwise stated.}
\label{tab:hyper}
\renewcommand{\arraystretch}{1}
\setlength{\tabcolsep}{1pt}
\resizebox{\linewidth}{!}{%
\begin{tabular}{lll}
\toprule
\textbf{Category} & \textbf{Hyperparameter} & \textbf{Value} \\
\midrule
\multirow{3}{*}{Weight quantization}
   & Bit-width $b_{W}$            & 4 \\
   & Scheme                       & Symmetric, per-group \\
   & Group size                   & 128 \\
\midrule
\multirow{2}{*}{Activation quantization}
   & Bit-width $b_{A}$            & 8 \\
   & Scheme                       & Symmetric, per-token \\
\midrule
\multirow{3}{*}{GPTQ solver}
   & Block size                   & 128 \\
   & Damping factor (percdamp)   & 0.01 \\
   & Activation order             & Disabled \\
\midrule
\multirow{3}{*}{Riemannian metric}
   & Sensitivity norm             & $\ell_{1}$ (Eq.~\eqref{eq:alpha_def}) \\
   & Weighting mode               & $\pi_{\ell}^{(m)}\!\propto\!\alpha_{\ell}^{(m)}$ \\
   & Covariance damping $\delta$  & $1\!\times\!10^{-4}$ \\
\midrule
\multirow{3}{*}{Calibration loss}
   & $\lambda_{W}$                & 1.0 \\
   & $\lambda_{A}$                & 1.0 \\
   & $\lambda_{\mathrm{geom}}$    & 0.2 \\
\midrule
Sparse rotation
   & Givens layers $K$            & 4 \\
\bottomrule
\end{tabular}
}
\vspace{-20pt}
\end{table}

Following recent work~\cite{li2025mbq,yu2025mquant}, we organize our comparisons into two complementary tiers and discuss them separately throughout this section.
(i) \emph{Classical Euclidean PTQ baselines}, including round-to-nearest (RTN), GPTQ~\cite{gptq}, AWQ~\cite{awq}, and SmoothQuant (SQ)\cite{smoothquant}. These methods are widely deployed in practical PTQ pipelines and provide a natural test bed for evaluating whether the unified Riemannian metric can improve Euclidean PTQ methods.
(ii) \emph{VLM-specific quantization baselines}, including Q-VLM, MBQ, MQuant, QuaRot\cite{quarot}, and SpinQuant~\cite{spinquant}. These methods cover representative multimodal quantization strategies, including modality-balanced reweighting, modality-specific static quantization, and dense rotation-based transformations.
For a fair comparison, all methods use group-wise quantization with group size $128$. For weight–activation joint quantization (W4A8), activations are quantized symmetrically per token, while weights are quantized symmetrically per output channel, enabling efficient execution on tensor cores. The default RGSQ hyperparameters are summarized in Table~\ref{tab:hyper}.

\textbf{Implementation details.}
All experiments are conducted on NVIDIA A100-80GB and A800-80GB GPUs.
The implementation is in Python 3.10 with PyTorch 2.9.0, CUDA 12.8, and Transformers 4.45.0.
Inference follows the default evaluation settings of LMMs-Eval~\footnote{\url{https://github.com/EvolvingLMMs-Lab/lmms-eval}}. 
The code is publicly available at RGSQ~\footnote{\url{https://github.com/RL-MIND/RGSQ}}.


\newcommand{\cmark}{\ding{51}}
\newcommand{\xmark}{\ding{55}}

\begin{table}[h]
\centering
\caption{Ablation study on LLaVA-OneVision-7B. \textbf{RMAM}: Modality-partitioned statistics with a unified Kronecker-Fisher metric. \textbf{ROW}: Riemannian objectives with whitening-based Euclidean reduction. \textbf{GGES}: Geometry-guided error steering via sparse Givens rotations.}
\small
{\renewcommand{\arraystretch}{0.9}
\resizebox{0.48\textwidth}{!}{
\setlength{\tabcolsep}{4pt}
\renewcommand{\arraystretch}{1.15}
\begin{tabular}{ccc|ccccccc}
\toprule
\multicolumn{3}{c|}{\textbf{Components}} &
\multirow{2}{*}{\textbf{MMMU}} &
\multirow{2}{*}{\textbf{SEED}} &
\multirow{2}{*}{\textbf{OCRB.}} &
\multirow{2}{*}{\textbf{VizWiz}} &
\multirow{2}{*}{\textbf{SciQA}} &
\multirow{2}{*}{\textbf{TextVQA}} &
\multirow{2}{*}{\textbf{Avg$\uparrow$}} \\
\cmidrule(r){1-3}
\textbf{RMAM} & \textbf{ROW} & \textbf{GGES} & & & & & & & \\
\midrule
\rowcolor{gray!15}
\multicolumn{3}{c|}{\textbf{FP16}} &
46.0 & 74.9 & 62.2 & 60.4 & 85.4 & 76.1 & 67.5 \\
\midrule

\xmark & \xmark & \xmark & 38.2 & 50.3 & 40.1 & 58.2 & 88.3 & 61.5 & 56.1 \\
\cmark & \xmark & \xmark & 44.8 & 66.5 & 54.2 & 59.5 & 88.8 & 70.8 & 64.1 \\
\cmark & \cmark & \xmark & 47.2 & 70.1 & 59.0 & 60.0 & 89.2 & 74.0 & 66.6 \\
\rowcolor{green!5}
\cmark & \cmark & \cmark & \textbf{49.2} & \textbf{72.4} & \textbf{62.3} & \textbf{60.4} & \textbf{90.0} & \textbf{76.0} & \textbf{68.4} \\
\bottomrule
\end{tabular}
}}
\label{tab:ablation}
\vspace{-10pt}
\end{table}

\subsection{Main Results}
\label{sec:exp_main}
\begin{table*}[h]
\centering
\caption{Main results on the large models of LLaVA-onevision, InternVL2, and Qwen2-VL families. $\dagger$ denotes our reported results.}
  {\renewcommand{\arraystretch}{0.95}
  \resizebox{0.9\textwidth}{!}{
      \begin{tabular}{cccccccccc}
        \toprule
        Model & Bitwidth & Method & MMMU & SEED & OCRBench & VizWiz & ScienceQA & TextVQA & Average ($\uparrow$) \\
        \midrule
        \multirow{10}{*}{LLaVA-onevision-72B} & FP16 & - & 56.1 & 78.1 & 73.2 & 69.2 & 90.0 & 79.3 & 74.3 \\
        \cmidrule(lr){2-10}
        ~ & \multirow{5}{*}{W3A16} & RTN & 53.9 & 77.4 & 68.2 & 66.1 & 89.5 & 77.4 & 72.1\\
        ~ & ~ & GPTQ & 52.7 & 76.0 & 69.7 & 68.3 & 89.3 & 77.9 & 72.3 \\
        ~ & ~ & AWQ & 33.4 & 71.2 & 48.7 & 49.3 & 69.2 & 58.8 & 55.1\\
        ~ & ~ & MBQ & 54.4 & 77.6 & 71.6 & 69.0 & 90.3 & 78.5 & 73.6\\
        ~ & ~ & RGSQ & \textbf{55.8} & \textbf{77.9} & \textbf{72.8} & \textbf{69.0} & \textbf{90.0} & \textbf{78.8} & \textbf{74.1}\\
        \cmidrule(lr){2-10}
        ~ & \multirow{4}{*}{W4A8} & RTN & 54.8 & 76.6 & 64.5 & 64.7 & 89.0 & 74.5 & 70.7\\
        ~ & ~ & SQ & 51.6 & 76.6 & 64.2 & 65.7 & 89.1 & 74.4 & 70.3\\
        ~ & ~ & MBQ & 55.6 & 76.5 & 64.4 & 65.7 & 89.2 & 73.3 & 70.8\\
        ~ & ~ & RGSQ & \textbf{56.0} & \textbf{78.0} & \textbf{73.0} & \textbf{69.0} & \textbf{90.0} & \textbf{79.2} & \textbf{74.2}\\
        \midrule
    
        \multirow{10}{*}{InternVL2-26B} & FP16 & - & 47.1 & 76.8 & 77.9 & 66.2 & 97.5 & 82.1 & 74.6\\
        \cmidrule(lr){2-10}
        ~ & \multirow{5}{*}{W3A16} & RTN & 46.6 & 75.7 & 75.9 & 64.7 & 96.4 & 80.6 & 73.3\\
        ~ & ~ & GPTQ & 44.8 & 75.8 & 76.0 & 60.9 & 96.3 & 80.1 & 72.3 \\
        ~ & ~ & AWQ & 46.4 & 76.2 & 76.4 & 64.5 & 96.7 & 81.0 & 73.5\\
        ~ & ~ & MBQ & 47.1 & 76.3 & 76.5 & 64.5 & 97.3 & 81.1 & 73.8\\
        ~ & ~ & RGSQ & \textbf{47.0} & \textbf{76.6} & \textbf{77.5} & \textbf{65.8} & \textbf{97.3} & \textbf{81.6} & \textbf{74.3}\\
        \cmidrule(lr){2-10}
        ~ & \multirow{4}{*}{W4A8} & RTN & 44.7 & 76.0 & 76.4 & 62.6 & 96.7 & 79.6 & 72.7\\
        ~ & ~ & SQ & 38.2 & 70.6 & 68.5 & 56.7 & 86.3 & 72.6 & 65.5\\
        ~ & ~ & MBQ & 44.0 & 75.7 & 77.5 & 62.0 & 97.1 & 80.0 & 72.7\\
        ~ & ~ & RGSQ & \textbf{47.6} & \textbf{76.8} & \textbf{77.8} & \textbf{66.0} & \textbf{97.4} & \textbf{81.8} & \textbf{74.6}\\

        \midrule
        
        \multirow{11}{*}{Qwen2-VL-72B} & FP16 & - & 61.1 & 77.6 & 79.9 & 76.0 & 91.6 & 82.5 & 78.1\\
        \cmidrule(lr){2-10}
        ~ & \multirow{5}{*}{W3A16} & RTN & 57.7 & 77.5 & 70.4 & 74.8 & 89.7 & 79.7 & 75.0\\
        ~ & ~ & GPTQ & 57.3 & 77.2 & 78.5 & 73.6 & \textbf{91.5} & 81.6 & 76.6 \\
        ~ & ~ & AWQ & 59.6 & 77.6 & 79.6 & 75.4 & 90.4 & 82.4 & 77.5\\
        ~ & ~ & MBQ & 59.6 & 77.7 & 79.4 & 75.6 & 90.5 & 82.5 & 77.6\\
        ~ & ~ & RGSQ & \textbf{60.8} & \textbf{77.6} & \textbf{79.8} & \textbf{75.8} & 91.2 & \textbf{82.4} & \textbf{77.9}\\
        \cmidrule(lr){2-10}
        ~ & \multirow{5}{*}{W4A8} & RTN & 58.1 & 76.6 & 66.2 & 71.3 & 90.1 & 77.0 & 73.2\\
        ~ & ~ & SQ & 55.9 & 76.4 & 65.5 & 69.7 & 88.8 & 76.9 & 72.2\\
        ~ & ~ & MBQ & 57.7 & 76.3 & 77.5 & 73.6 & 89.6 & 80.5 & 75.8\\
        ~ & ~ & MQuant$^\dagger$ & 59.8 & 76.5 & 78.6 & 73.5 & 87.8 & 81.6 & 76.3\\
        ~ & ~ & RGSQ & \textbf{61.0} & \textbf{77.5} & \textbf{80.0} & \textbf{75.8} & \textbf{91.4} & \textbf{82.4} & \textbf{78.0}\\
        
        \bottomrule
      \end{tabular}
    }}
  \label{tab:main-large}
  \vspace{-10pt}
\end{table*}

\textbf{Weight-only quantization (W3A16).}
As shown in Table~\ref{tab:main-small} and Table~\ref{tab:main-large}, RTN reveals that smaller VLMs are substantially more sensitive to weight quantization than larger ones: averaged over the three small models, RTN drops by 10.2\% from FP16, whereas the drop on the three large models is only 2.2\%. This suggests that limited capacity amplifies quantization noise once errors propagate through a mixed-modality token interface. Under W3A16, RGSQ consistently outperforms the modality-agnostic baselines RTN, GPTQ, and AWQ across all six models. Within the LLaVA-OneVision family alone, RGSQ improves the average score over AWQ by $+6.9\%$ on the 7B model and by $+19.0\%$ on the 72B model. Particularly striking is LLaVA-OneVision-72B, where AWQ collapses by $-17.0\%$ relative to RTN, while RGSQ restores performance to $74.1\%$, essentially matching the $74.3\%$ FP16. These results corroborate our central claim: explicit modeling of modality-dependent sensitivity is necessary for reliable quantization in VLMs.

\textbf{Weight–activation quantization (W4A8).}
The same trend holds, with an even larger margin under joint quantization. RGSQ consistently outperforms SmoothQuant and RTN under W4A8, with gains reaching $+17.0\%$ in representative cases (Tables~\ref{tab:main-small}, \ref{tab:main-large}). We also observe that SmoothQuant is often inferior to plain RTN on InternVL2-26B, it drops from $72.7\%$ to $65.5\%$, indicating that naive activation re-scaling can exacerbate inter-modality imbalance once activations are quantized. RGSQ, in contrast, achieves stable improvements on both small and large models: it raises InternVL2-26B to $74.5\%$ and Qwen2-VL-72B to $78.0\%$, virtually indistinguishable from the $78.1\%$ FP16. Overall, W4A8 results further confirm that multimodal-aware quantization is essential, especially once activation quantization makes modality sensitivity disparities more pronounced.

\subsection{Enhancing Baseline Quantizers via ROW}
\label{sec:exp_row}

\begin{table*}[h]
\centering
\caption{Effect of our Riemannian whitening on existing quantization methods for InternVL2-8B. This geometry-aligned method converts the Riemannian objective into equivalent Euclidean reconstruction in whitened coordinates, enabling plug-and-play enhancements to standard PTQ solvers during calibration without modality-specific scaling.}
  {\renewcommand{\arraystretch}{0.95}
  \resizebox{0.9\textwidth}{!}{
      \begin{tabular}{cccccccccc}
        \toprule
        Model & Bitwidth & Method & MMMU & SEED & OCRBench & VizWiz & ScienceQA & TextVQA & Average ($\uparrow$) \\
        \midrule
        \multirow{7}{*}{InternVL2-8B} & FP16 & - & 48.0 & 76.0 & 76.5 & 61.1 & 96.2 & 77.0 & 72.5\\
        \cmidrule(lr){2-10}
        ~ & \multirow{6}{*}{W4A8} & RTN & 44.3 & 74.0 & 72.0 & 57.1 & 95.5 & 73.1 & 69.3\\
        ~ & ~ & \quad w/ WER & 46.2 & 75.0 & 74.5 & 58.8 & 95.9 & 75.0 & 70.9 \textcolor{blue}{\scriptsize(+1.6)}\\
        ~ & ~ & AWQ & 41.8 & 73.8 & 70.9 & 54.2 & 95.1 & 72.6 & 68.1\\
        ~ & ~ & \quad w/ WER & 44.5 & 74.8 & 73.8 & 57.0 & 95.6 & 74.5 & 70.0 \textcolor{blue}{\scriptsize(+1.9)}\\
        ~ & ~ & SQ & 45.6 & 74.3 & 73.0 & 56.5 & 95.8 & 72.3 & 69.6\\
        ~ & ~ & \quad w/ WER & \textbf{47.0} & \textbf{75.4} & \textbf{75.2} & \textbf{59.2} & \textbf{96.0} & \textbf{75.6} & \textbf{71.4} \textcolor{blue}{\scriptsize(+1.8)}\\
        \bottomrule
      \end{tabular}
    }}
  \label{tab:white}
  \vspace{-10pt}
\end{table*}
A direct corollary of Eq.~\eqref{eq:euclid_reduced} is that any Euclidean block-wise PTQ method can be wrapped with our Riemannian whitening (ROW) to inherit modality-aware curvature scoring at calibration time, while leaving the runtime computation graph unchanged.
We verify this claim on InternVL2-8B under W4A8 (Table~\ref{tab:white}).
The vanilla RTN, AWQ, and SmoothQuant baselines incur $2.9$–$3.2\%$ average degradations against FP16, with particularly severe losses on vision-reliant tasks such as VizWiz.
Adding ROW yields a consistent uplift of $+1.6\%$ to $+1.9\%$ across all three baselines without altering their core quantization logic, closing the FP16 gap to within $1.1\%$. These results show that the Riemannian whitening reduction is a broadly applicable enhancement, not a single-method trick.

\subsection{Low-Bit Quantization for VLMs}
\label{sec:exp_lowbit}
\begin{figure*}[t]
    \centering
    \includegraphics[width=0.9\linewidth]{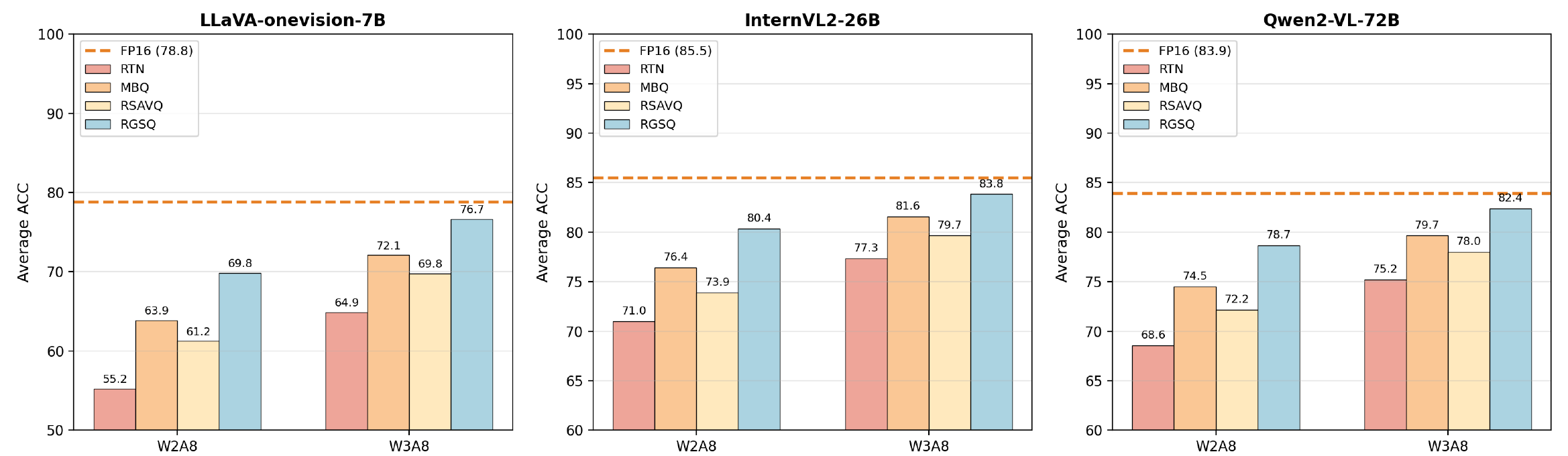}
    \caption{Average performance across three benchmarks (ScienceQA, SEED, and TextVQA) under low-bit quantization settings (W2A8 and W3A8). RSAVQ is designed for LLMs and shows limited gains on VLMs, while MBQ and RGSQ are VLM-aware methods.}
    \label{fig:lowbit}
    \vspace{-5pt}
\end{figure*}

\begin{figure*}[t]
    \centering
    \includegraphics[width=0.9\linewidth]{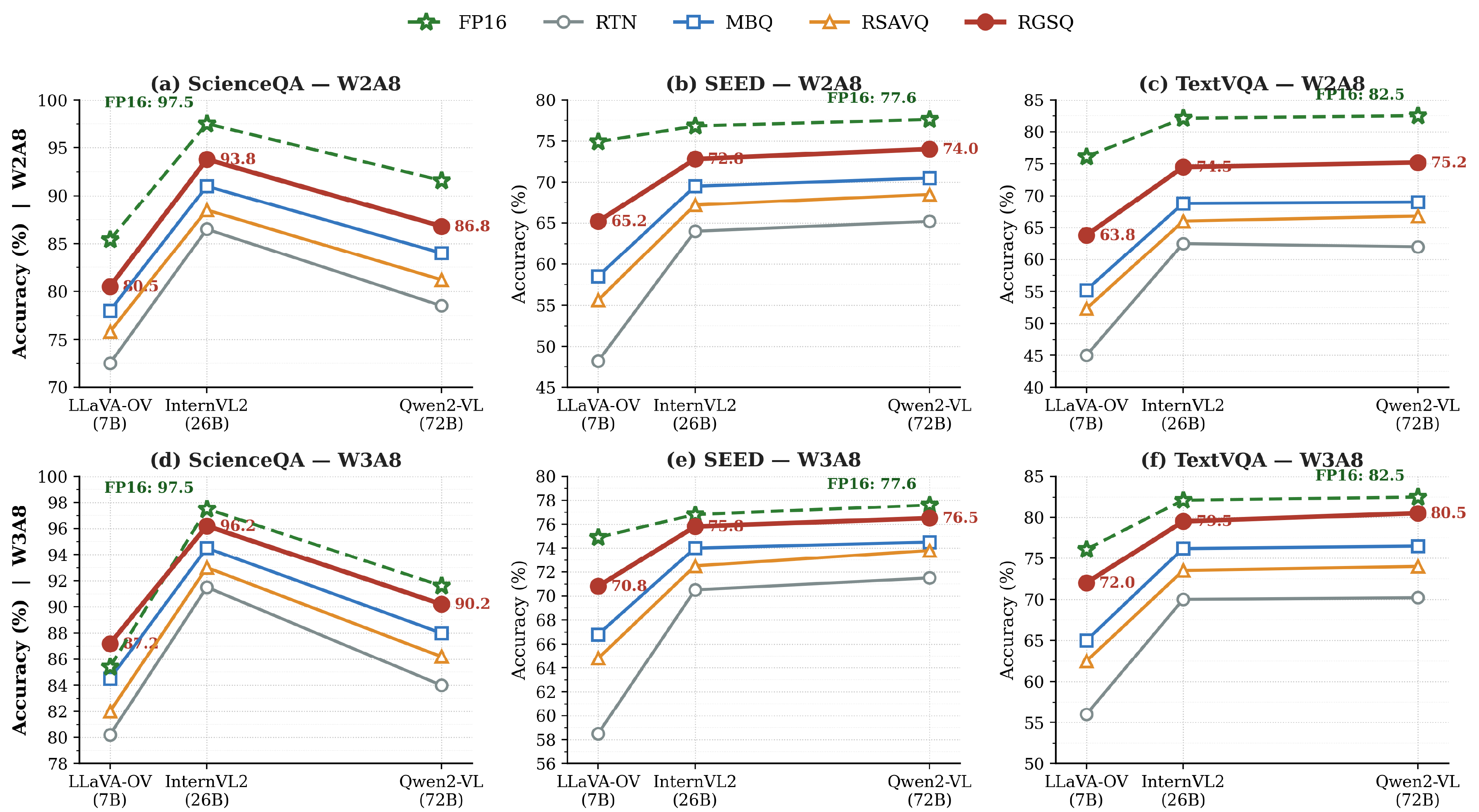}
    \caption{Low-bit quantization results under W2A8 and W3A8 settings. RSAVQ is designed for LLMs and shows limited gains on MLLMs, while MBQ and RGSQ are MLLM-aware methods.}
    \label{fig:lowbit_6p}
    \vspace{-5pt}
\end{figure*}

We further evaluate the more aggressive W2A8 and W3A8 settings, where the assumptions of Euclidean PTQ become increasingly fragile. Under such extreme bit widths, RTN suffers substantial accuracy degradation, especially on vision-intensive benchmarks, as shown in Fig.\ref{fig:lowbit}. MBQ remains limited despite its VLM-aware design, while RSAVQ has only a modest effect in this setting because it was originally developed for LLM quantization rather than VLMs.
As shown in Fig.\ref{fig:lowbit_6p}, RGSQ improves over RTN by up to $+14.6\%$ under W2A8 and consistently outperforms both MBQ and RSAVQ. Under W3A8, RGSQ reduces the average gap to FP16 to only $1.5$--$2.1\%$.
This advantage becomes more evident on larger models. Table~\ref{tab:large_lowbit} shows that RGSQ outperforms MBQ by $+4.0\%$ to $+4.2\%$ under W2A8 on Qwen2-72B and InternVL2-26B, where Euclidean reconstruction and scalar reweighting are particularly limited. These results are consistent with the motivation of the Riemannian formulation. As bit widths decrease, the direction of quantization error has a stronger effect on task loss, and directional curvature information becomes essential for redistributing the quantization budget across modalities.

\begin{table}[t]
\centering
\small
\caption{Low-bit quantization results on large VLMs. We report the 6-benchmark average accuracy under W2A8 and W3A8 settings.}
\label{tab:large_lowbit}
\renewcommand{\arraystretch}{1.05}
\setlength{\tabcolsep}{4pt}
\resizebox{\linewidth}{!}{%
\begin{tabular}{llccccc}
\toprule
\textbf{Model} & \textbf{Setting} & \textbf{RTN} & \textbf{MBQ} & \textbf{RGSQ} & \textbf{$\Delta$ vs.\ MBQ} \\
\midrule
\multirow{2}{*}{Qwen2-VL-72B}
   & W2A8 (avg) & 68.6 & 74.5 & \textbf{78.7} & \textcolor{blue}{+4.2} \\
   & W3A8 (avg) & 75.2 & 79.7 & \textbf{82.4} & \textcolor{blue}{+2.7} \\
\midrule
\multirow{2}{*}{InternVL2-26B}
   & W2A8 (avg) & 71.0 & 76.4 & \textbf{80.4} & \textcolor{blue}{+4.0} \\
   & W3A8 (avg) & 77.3 & 81.6 & \textbf{83.8} & \textcolor{blue}{+2.2} \\
\bottomrule
\end{tabular}}
\end{table}

\subsection{Ablation Study}
\label{sec:exp_ablation}
We isolate the contribution of each RGSQ component on LLaVA-OneVision-7B under W4A8, with the modality-agnostic Euclidean PTQ as the starting point (Table~\ref{tab:ablation}). The baseline yields an average accuracy of $56.1\%$, a substantial $-11.4\%$ drop relative to the $67.5\%$ FP16. Adding the Riemannian Modality-Aware Metric (\textbf{RMAM}), which builds a unified Kronecker–Fisher metric from modality-partitioned statistics, recovers $+8.0\%$ to $64.1\%$, confirming that modality-specific curvature estimation is the key to stable multimodal calibration. Layering on Riemannian Objectives with Whitening (\textbf{ROW}) further raises the average to $66.6\%$, showing that geometry-aligned reconstruction outperforms isotropic Euclidean objectives once the metric is in place. Finally, activating Geometry-Guided Error Steering (\textbf{GGES}) via sparse Givens rotations brings the model to $68.4\%$, slightly surpassing the FP16 baseline.

\subsection{Generalization to Video Understanding}
\label{sec:exp_video}
\begin{table}[t]
\centering
\small
\caption{Generalization to video understanding. RGSQ consistently improves over MBQ on SEED-Image, SEED-Video, and Video-MME.}
\label{tab:video}
\renewcommand{\arraystretch}{1.05}
\setlength{\tabcolsep}{4pt}
\resizebox{\linewidth}{!}{%
\begin{tabular}{llccc}
\toprule
\textbf{Model} & \textbf{Method} & \textbf{SEED-Image} & \textbf{SEED-Video} & \textbf{Video-MME} \\
\midrule
\multirow{3}{*}{InternVL2-8B}
    & FP16  & 76.2 & --   & 56.9 \\
    & MBQ   & 74.3 & 50.8 & 51.2 \\
    & \textbf{RGSQ} & \textbf{76.0} & \textbf{52.4} & \textbf{54.3} \\
\midrule
\multirow{3}{*}{LLaVA-OneVision-7B}
    & FP16  & 74.9 & 56.9 & 58.2 \\
    & MBQ   & 69.7 & 50.7 & 54.0 \\
    & \textbf{RGSQ} & \textbf{72.4} & \textbf{53.5} & \textbf{56.9} \\
\bottomrule
\end{tabular}}
\vspace{-10pt}
\end{table}

Although RGSQ is calibrated on a small set of static image-text pairs and is not designed with video models in mind, the learned modality-aware Riemannian metric may remain informative for video tokens, since these tokens inherit the same visual-versus-text statistical asymmetry observed in the image setting. As a preliminary check on this hypothesis, we directly evaluate the W4A8-quantized models on the SEED-Image, SEED-Video, and Video-MME benchmarks under the LMMs-Eval video protocols, without performing any additional video calibration.
As reported in Table~\ref{tab:video}, RGSQ improves over the MBQ baseline across all three benchmarks for both InternVL2-8B and LLaVA-OneVision-7B, with the video-specific gains on SEED-Video and Video-MME ranging from $1.6\%$ to $3.1\%$. These early results suggest that the unified Riemannian metric retains part of its benefit in temporal multimodal settings and does not appear to be tightly bound to static image distributions.
These experiments are intended as an initial verification. RGSQ is developed as a general-purpose VLM quantization framework, so the observed video gains are better read as a sign of generalization than as evidence of video-specialized performance. A more thorough evaluation with video-specific calibration and a broader set of video benchmarks is left to future work.

\subsection{Layer-Wise Geometric Diagnostics}
\label{sec:exp_geom_diag}
\begin{figure*}[t]
\centering
\includegraphics[width=0.86\linewidth]{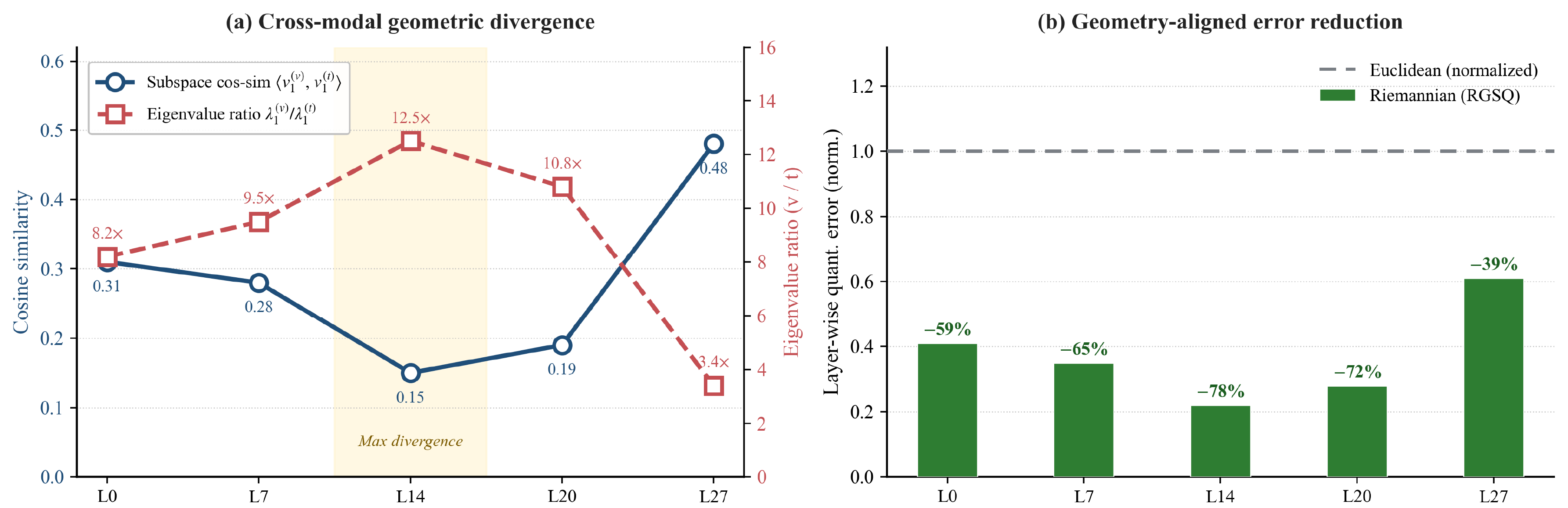}
\caption{Layer-wise geometric diagnostics on LLaVA-OneVision-7B under W4A8 quantization. We report modality eigenspace divergence and the corresponding reduction in normalized quantization error.}
\label{fig:layerwise}
\vspace{-5pt}
\end{figure*}

To understand the source of these improvements, we probe the layer-wise geometry of LLaVA-OneVision-7B under W4A8 quantization (Fig.~\ref{fig:layerwise}). At five representative layers, we measure the cosine similarity between the leading eigenvectors of the modality-partitioned activation covariances $A_{\ell}^{(v)}$ and $A_{\ell}^{(t)}$, and the leading eigenvalue ratio $\lambda_{1}^{(v)}/\lambda_{1}^{(t)}$. Lower cosine similarity indicates more orthogonal modality subspaces, while a higher eigenvalue ratio indicates stronger energy asymmetry.

In the mid-fusion block (L14), the cosine similarity drops to $0.15$, while the eigenvalue ratio rises to $12.5\times$. This suggests that vision and text tokens occupy nearly orthogonal subspaces with highly asymmetric energy, which cannot be captured by a single per-modality scalar.

Replacing the Euclidean error proxy with the Riemannian metric in Eq.~\eqref{eq:riem_weight_obj} reduces the normalized quantization error by up to $78\%$ at the layers with the largest modality divergence. The reduction is smaller in the generation-heavy tail (L27), where the two modalities are already more aligned. This layer-wise correspondence supports the use of directional curvature information rather than scalar reweighting in RGSQ.

\subsection{Calibration Efficiency Analysis}
\label{sec:exp_calib_eff}

\begin{table}[t]\small
\centering
\caption{Comparison of calibration cost and accuracy across different quantization methods.}
\label{tab:calib_cost}
\resizebox{\linewidth}{!}{
\begin{tabular}{lccccc}
\hline
\textbf{Method} & \textbf{Time} & \textbf{Mem.} & \textbf{GPU-h} & \textbf{Avg. Acc.} & $\Delta$ \textbf{FP16} \\
\hline

GPTQ &  35 min & 22 GB & 0.58 & 64.1\% & $-3.4$ \\

AWQ &  25 min & 20 GB & 0.42 & 60.6\% & $-6.9$ \\

MBQ &  45 min & 24 GB & 0.75 & 65.8\% & $-1.7$ \\

MQuant &  1.2 h & 28 GB & 1.2 & 66.6\% & $-0.9$ \\

RGSQ (Ours) &  2.1 h & 38 GB & 2.1 & 68.1\% & $+0.6$ \\

\hline
\end{tabular}
}
\end{table}
Table~\ref{tab:calib_cost} quantifies the practical cost of RGSQ relative to lightweight PTQ baselines. On LLaVA-OneVision-7B at W4A8, the full pipeline requires approximately $2.1$\,GPU-hours and $38$\,GB of peak memory on a single A800, compared with $0.42$--$1.2$\,GPU-hours for GPTQ, AWQ, MBQ, and MQuant. The additional overhead stems from two sources: estimation of the modality-partitioned activation and gradient second-order factors, which is more expensive than purely feed-forward calibrators such as AWQ, and the sparse Givens search, which introduces a structured discrete optimization step that nonetheless remains cheaper and far more parameter-efficient than unconstrained dense rotation learning. Because the whitening step is applied only at calibration time and is absorbed into the quantized parameters before deployment, RGSQ introduces no extra runtime operator. Despite the moderate increase in offline cost, RGSQ achieves superior performance and the highest accuracy among all compared methods, surpassing even the FP16 baseline.
\section{Conclusion and Limitation}
\label{sec:conclusion}

In this paper, we introduced \textbf{RGSQ}, a geometry-sensitive PTQ framework for vision–language models. Our experiments reveal a central observation: the quantization error of VLMs is governed not only by its magnitude, but also by its modality-dependent geometric direction. Vision and text tokens often induce distinct layer-wise curvature and sensitivity patterns, making conventional Euclidean reconstruction, scalar reweighting, or modality-agnostic rotations insufficient for reliable low-bit quantization. RGSQ addresses this issue by measuring quantization error on a unified modality-aware Riemannian manifold and steering the error toward low-sensitivity directions. This geometric treatment consistently improves weight-only and weight–activation quantization across model families, scales, and bit-widths, and remains effective when used as a plug-in enhancement for existing Euclidean PTQ solvers. The gains are most pronounced under aggressive low-bit settings and at layers with strong vision–text geometric divergence, suggesting that modality-aware curvature is a key factor for robust VLM compression. RGSQ introduces a moderate one-time calibration overhead from layer-wise sensitivity estimation and sparse rotation search. A practical future direction is to improve the efficiency of geometry-aware calibration through faster Fisher estimation, lighter rotation search, and finer token-level sensitivity modeling.

\bibliographystyle{IEEEtran}
\bibliography{RGSQ}

\vfill
\end{document}